\documentclass[preprint,12pt]{elsarticle}

\usepackage{amssymb}
\usepackage{amsmath}
\usepackage{subfig}

\journal{Bioinspiration \& Biomimetics}

\begin{document}

\begin{frontmatter}



\title{RoboANKLE: Design, Development, and Functional Evaluation of a Robotic Ankle with a Motorized Compliant Unit}


\affiliation[label1]{organization={Human-Centered Design Laboratory,
Department of Mechanical Engineering,
Ozyegin University},
            city={Istanbul},
            country={Turkey}}
\author{Baris Baysal}
\author{Omid Arfaie} 
\author{Ramazan Unal\corref{cor1}}
\cortext[cor1]{Corresponding Author}
\ead{ramazan.unal@ozyegin.edu.tr}

\begin{abstract}
This study presents a powered transtibial prosthesis with complete push-off assistance, RoboANKLE. The design aims to fulfill specific requirements, such as a sufficient range of motion (RoM)  while providing the necessary torque for achieving natural ankle motion in daily activities. Addressing the challenges faced in designing active transtibial prostheses, such as maintaining energetic autonomy and minimizing weight, is vital for the study. With this aim, we try to imitate the human ankle by providing extensive push-off assistance to achieve a natural-like torque profile. Thus, Energy Store and Extended Release mechanism (ESER) is employed with a novel Extra Energy Storage (EES) mechanism. Kinematic and kinetic analyses are carried out to determine the design parameters and assess the design performance. Subsequently, a Computer-Aided Design (CAD) model is built and used in comprehensive dynamic and structural analyses. These analyses are used for the design performance evaluation and determine the forces and torques applied to the prosthesis, which aids in optimizing the design for minimal weight via structural analysis and topology optimization. The design of the prototype is then finalized and manufactured for experimental evaluation to validate the design and functionality. The prototype is realized with a mass of 1.92 kg and dimensions of 261x107x420 mm. The Functional evaluations of the RoboANKLE revealed that it is capable of achieving the natural maximum dorsi-flexion angle with 95\% accuracy. Also, Thanks to the implemented mechanisms, the results show that RoboANKLE can generate 57\% higher than the required torque for natural walking. The result of the power generation capacity of the RoboANKLE is 10\% more than the natural power during the gait cycle. 

\end{abstract}


\end{frontmatter}



\section{Introduction}

In 2005, there were 1.6 million recorded lower limb amputees in the United States, with a forecasted increase to 3.6 million by 2050 \cite{ZieglerGraham2008}.

Activities of daily living (ADL) is indicates person's functional status. After lower limb amputations, users struggle to perform ADL tasks \cite{DeRosendeCeleiro2016}. Prosthesis are devices that is designed to support individuals to perform ambulatory ADL. In the literature, lower limb prostheses are classified for the power intake types as passive, active, and semi-active  prostheses \cite{Weerakkody2017}. Passive prostheses do not incorporate external actuators where there is no power intake, requiring the amputee’s residual muscles to compensate the necessary torque at the ankle joint to facilitate walking  \cite{Winter1988}. Widely available passive prostheses are found in two different types: Solid Ankle Cushioned Heel (SACH) and Energy Storing And Releasing foot (ESAR). SACH foot has a rigid internal structure and visual appearance of a foot. Designed to provide a walking functionality in a basic sense and standing support. On the other hand, ESAR foot are passive prostheses with several mechanisms to store energy and deliver it to the foot during walking for reducing the metabolic cost of walking \cite{Chiriac2020,Ehara1993}. Net positive work cannot be generated via passive prosthesis \cite{Hunt2023}, due to extra metabolic energy consumed. Conversely, active prostheses have ability to provide net positive work \cite{Herr2011}, since they are externally actuated devices with a positive power intake. These prostheses, in general generate net positive work torque for the ankle joint \cite{Au2007}. Semi-active prostheses (hybrid), however, utilizes relatively small actuators compared to the fully active prosthesis for generating net positive work, thanks to their energy storing elements which would result in decreased mass and mechanical energy consumption \cite{Glanzer2018},\cite{Adamczyk2020}.

Metabolic energy consumption is an essential issue in lower limb prosthetics. Prosthesis users spends more metabolic energy even in walking compared to able-bodied persons \cite{Czerniecki1996, Genin2008, Waters1976}. On average amputees spends 30\% more metabolic energy during walking with transitibial prostheses \cite{Asif2021}. Oxygen consumption also increases with 20\% while walking at normal walking speed \cite{Molen1973}, moreover, acceleration in walking increases oxygen consumption along with heart rate \cite{vanSchaik2019}. Powered ankle–foot prostheses are shown to reduce metabolic demand and enhance step-to-step transition work during level walking when compared to passive devices \cite{esposito2016}.
In passive prostheses, shown that ESAR foot is able produce more work during push-off that shows the mechanical advantage \cite{Houdijk2014}. Thus, resulting in lower energy expenditure with ESAR prostheses \cite{Hsu2006}. In this study, address this issue, proposed prosthesis is employing dual ESAR mechanism with one of them incorporating an active element.

The weight of the prosthesis is another essential specification that need to be considered while designing a transtibial prosthesis. The excessive mass of the prostheses is one of the main reasons that may lead to reject the prosthesis by user \cite{Gailey2010}. In the active prosthesis, increment of the active elements leads the increased bulkiness and mass of the prostheses \cite{Weerakkody2017}. Another factor leading excessive mass in prosthesis is having battery with a sizable mass. High power intake of the actuators would require more energy which resulting bigger and heavier batteries. Choosing the correct battery is essential to ensure that it is sufficient for the user entire day ADL tasks without depleting, although it should be small to avoid excessive mass. To ensure having a lightweight battery, the energy efficiency of the actuators is required. However, adding more active components would result in a heavier prosthesis, increasing the overall bulk of the system and causing discomfort for the amputee when used for prolonged sessions \cite{Prentice2004}. In other words, excessive weight leads to discomfort by placing additional strain on the connection between the socket and the limb \cite{Sewell2000}. Furthermore, amputees experience walking asymmetries and decreased mobility as a result of socket-related pain \cite{Hak2014}. Efforts to minimize prosthesis weight have incorporated the ESAR mechanism, which has also demonstrated a reduction in metabolic cost for amputees \cite{Gao2016,Lecomte2021}. Incorporating ESAR mechanisms into prosthetic systems enhances efficiency by reducing the instantaneous power demands on actuators and lowering overall battery consumption compared to fully direct-drive actuated systems \cite{laschowski2023, lemoyne2016}. ESAR-based prostheses, such as those employing series elastic actuators, are shown to recycle mechanical energy during gait, enabling smaller battery packs and extending operational time without sacrificing performance \cite{mazzarini2023, ventura2011}. However, a robotic ankle prosthesis that is able to meet the requirements of the ADL and at the same time be as light as possible to reduce the metabolic energy consumption by the user is still challenges in this field.

In this study, two main energy generator units are proposed to meet the requirement of the fast walking within the robotic ankle prosthesis. The DF mechanism stores the energy starting from the mid-stance phase  till the push-off phase and release that energy at the push-off pahse. The EES mechanism employs a small motorized spring to compensate the remaining part of the required torque around the ankle joint for the desired task. The kinematic and kinetic simulations are performed and based on the achieved forces and torques on the different components of the RoboANKLE the, the structural analyses are carried out. The RoboANKLE is manufactured mainly from Al 7075 material. The shafts and pins are manufactured from stainless steel. The overall mass of the RoboANKLE is measured as 1.92 kg with dimensions as $26\times107\times420 mm$. To assess the capacity of the torque and power generating of the RoboANKLE, the characterization of it is performed. The characterization results, the RoboANKLE is able to produce the desired torque and power 13\% faster and deliver 7.7\% more torque comparing to the biomechanical data \cite{winter}. In order to evaluate the functionality of the RoboANKLE, the treadmill walking with and able-bodied participant is performed. Using the XSENS the kinematic measurements are done. The functional evaluation results of the RoboANKLE revealed that, thanks to the implemented mechanisms, it generates 57\% , 10\% higher torque and power compared to the biomechanical data.

\section{Design of RoboANKLE}

In this section, the design and development of the RoboANKLE are illustrated. The conceptual design of the RoboANKLE, illustrating the mechanisms utilized in this prosthesis and their specification, are described. Then, the detailed CAD model of the RoboANKLE is developed. According to the proposed conceptual design, the kinematic and kinetic analyses of the RoboANKLE are conducted. The selection of the proper actuator system for the ankle joint based on the requirement of the ankle prosthesis is proposed. Additionally, the structural analyses of the RoboANKLE components are conducted according to joint forces and torques from the kinematic analyses results.

\subsection{Mechanisms Design \& and CAD Modelling}

In order to deliver the required torque and power of the walking, the RoboANKLE utilizes two ESAR mechanisms. The first one is the Dorsi-Flexion (DF) mechanism, storing the energy during the mid-stance of the walking and release it at the push-off stage. The remaining of the required torque and power of the task is compensating with the second mechanism named Extra Energy Store (EES) mechanism. This mechanism employs a motorized spring actuator, injecting the torque to the ankle joint at the push-off phase. It is hypothesis that using these two mechanisms, the total required torque and power for the fast walking of a 75-kg participant is realized.

The concept of the DF mechanism as illustrated in our previous study \cite{BaysalUnal2024}, is based on storing energy during the dorsi-flexison of the ankle joint  from the mid-stance phase to the starting instant of the push-off phase. The stored energy during this period is delivered to the ankle joint during the push-off phase. To make the torque profile of the push-off phase in the RoboANKLE closer to the human natural profile, the DF springs are compressed during mid-stance to dorsi-flexison maximum. Then the stored energy during the dorsi-flexison (within a $10^{\circ}$ range) is released during plantar-flexion (within a $25-30^{\circ}°$ range). 

\begin{figure}[h]
	\centering
	\includegraphics[width=0.45\textwidth]{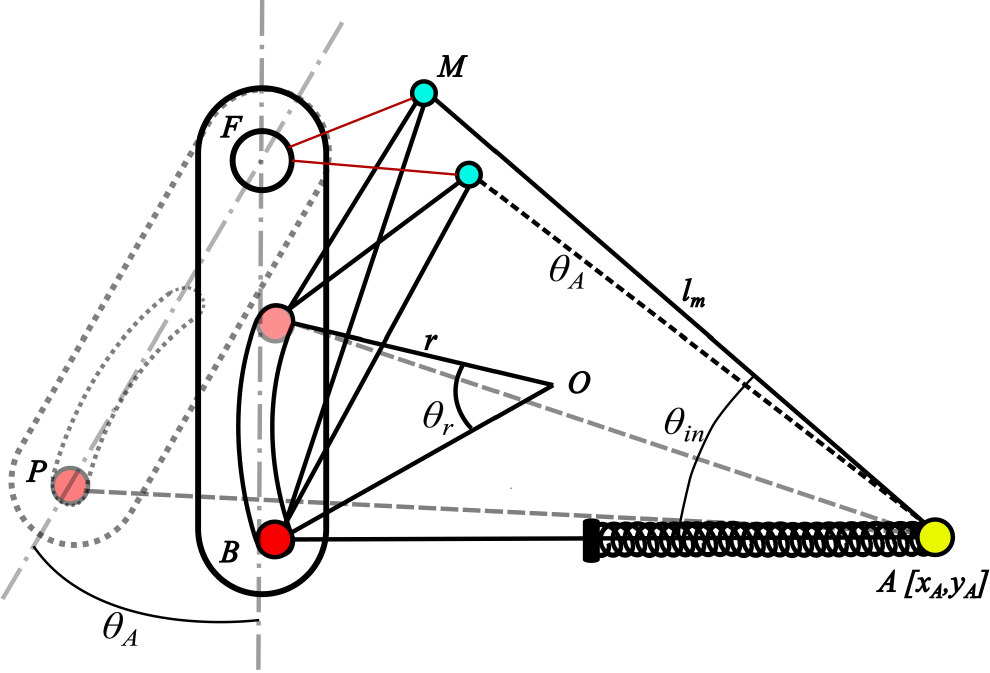}
	\caption{The schematic model of the DF mechanism}
	\label{dorsi_mechanism}
\end{figure}

\begin{figure}[h]
	\centering
	\hspace{-8mm}
	\subfloat[]{
		\includegraphics[width=0.27\textwidth]{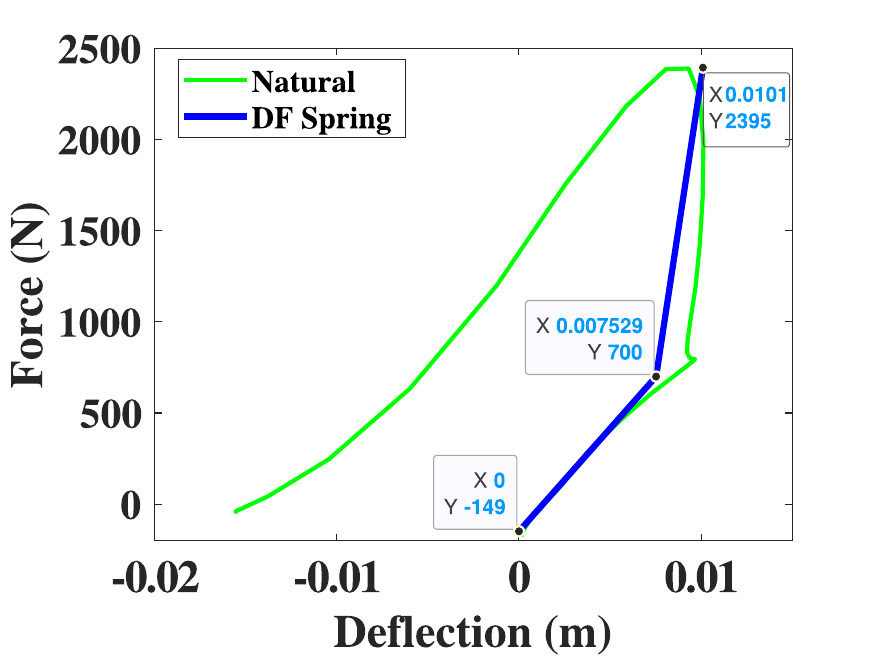}
		\label{subfig:force_deflection_relationship}
	}
	\hspace{-8mm} 
	\subfloat[\scriptsize]{
		\includegraphics[width=0.27\textwidth]{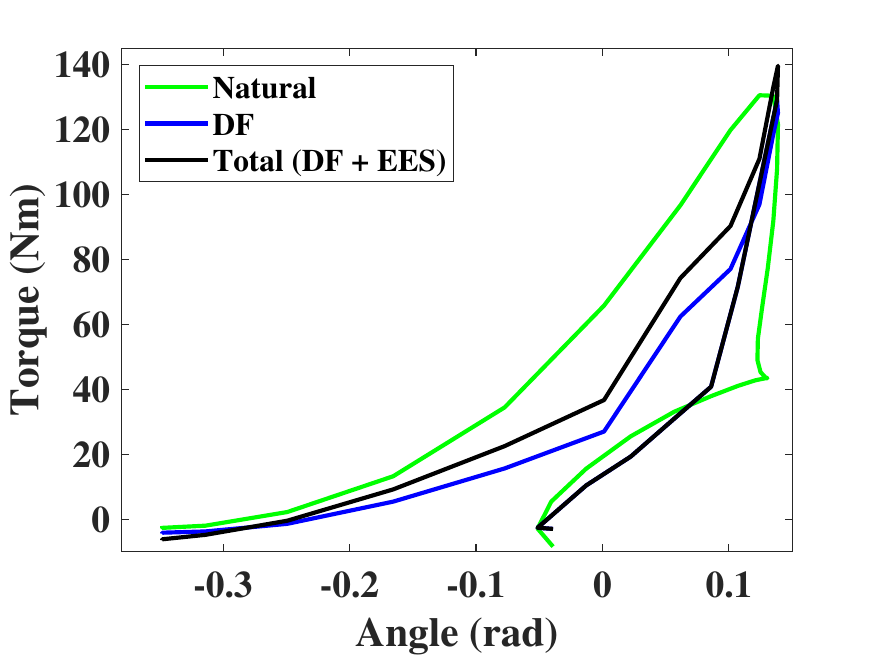}
		\label{subfig:df_ees_torque}
	}
	\hspace{-8mm}
	\subfloat[\scriptsize]{
		\includegraphics[width=0.27\textwidth]{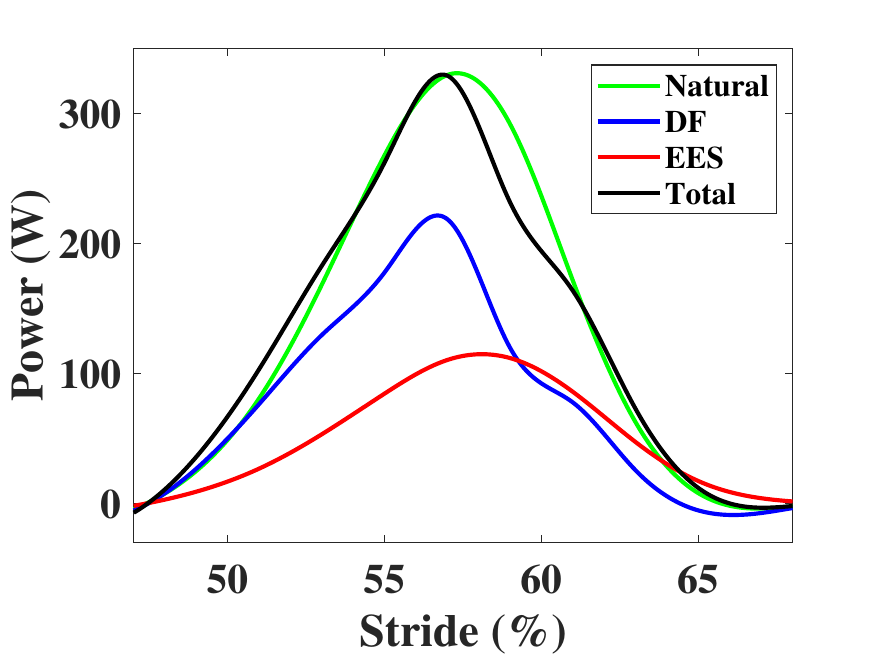}
		\label{subfig:df_ees_power_generation}
	} 
	\hspace{-8mm} 
	\subfloat[\scriptsize]{
		\includegraphics[width=0.27\textwidth]{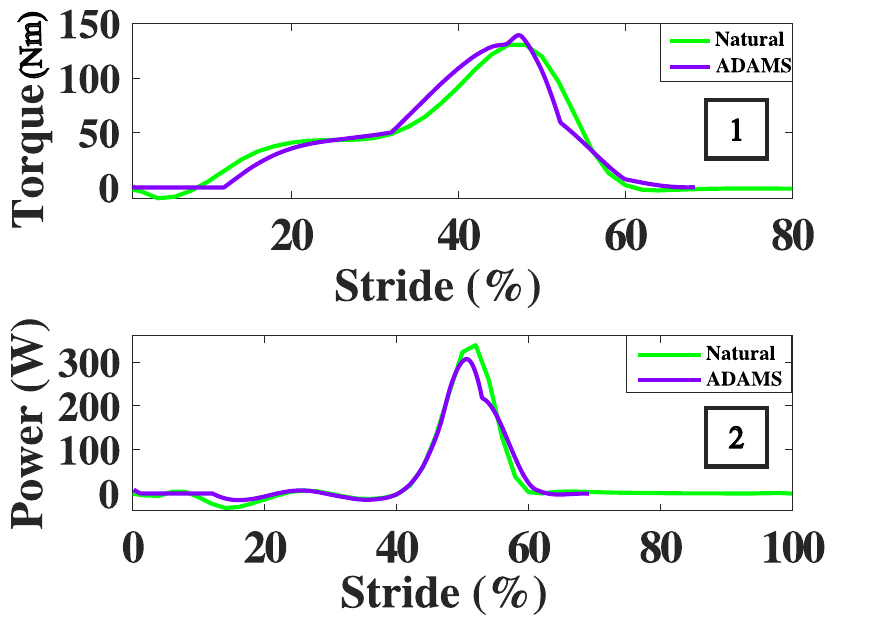}
		\label{subfig:adams_torque_power}
	}
	
	\caption{Comparison of the natural ankle rotation deflection with linear spring compression within the force generated by the ankle (a), comparison between the natural ankle angle-torque profile (green) and generated torque by the DF mechanism (black) and the total generated torque around the ankle joint (b), comparison between the power flow around the natural ankle (green), the generated power profile by the DF mechanism (green), EES mechanism (blue), and the total power profile of the RoboANKLE (black) (c), the Adams simulation results for the generated ankle torque and natural torque during walking (d-top), Adams simulation ankle power flow and natural power flow during walking (d-bottom)}
	\label{fig:ankle_eval}
\end{figure}

Using the relation between the natural ankle angle and the ankle torque allows us to approximate the joint stiffness during the dorsi-flexison phase. To determine the stiffness of the DF springs, a predefined properties for the moment arm length and initial location is selected. These predefined properties of the moment arm align with the maximum accessible length and location in the design. According to the selected properties of the moment arm, the force-deflection relation of the ankle joint is derived as shown in Fig. \ref{subfig:force_deflection_relationship}. To achieve a non-linear torque profile, two parallel springs with different lengths and spring constants are selected to mimic the force-deflection profile.
The DF springs attachment point is connected to a sliding joint which rotates in a sliding path as shown in Fig. \ref{dorsi_mechanism}. The sliding path is designed in an arc shape to ensure that the length of the spring remains constant in maximum dorsi-flexison where springs are compressed fully. Consequently, as the springs attachment point moves inside the sliding path, no energy will loss during the energy releasing at the push-off phase. The final position of the slider path is set based on the toe-off plantar-flexion angle. At this point all the stored energy in the spring will be fully released. In another word, the deflection of the DF springs will be zero at the toe-off phase. Using this concept, the natural push-off of the ankle joint will mimic through the releasing of energy from the springs throughout the push-off phase. The properties of the sliding path such as the radius, starting, and end point of it can be adjusted for different users with different maximum dorsi-flexison angle and  various walking speeds.

In Fig. \ref{dorsi_mechanism}, the schematic of the DF mechanism is shown. 
In this figure, the point $A$ represents DF springs attachment point at the toe and point $F$ shows the ankle joint. Point $B$ is the initial position of the spring attachment inside the sliding path at the mid-stance phase and point $P$ represents the position of the attachment point inside the sliding path at the starting instance of the push-off phase. This rotation is due to the shank's rotation around the ankle joint. $O$ is the  arc center of the sliding path and $r$ is the radius of this arc. The deflection of the DF spring, is determined through the calculation of the length of $AP$ as shown in equations (\ref{FM}-\ref{AP}).
 
The \(\overrightarrow{FM}\) vector is calculated using Equation \eqref{FM}, where \(F\) denotes the location of the ankle joint, $M$ represents replacement rotation point, $m$ is the length of \(\overrightarrow{FM}\) vector which is constant, and $\theta_{A}$ is the ankle joint rotation. 

\begin{equation}{\label{FM}}
	\overrightarrow{FM}=-m*\cos{\theta_{A}}\overrightarrow{i}+m*\sin{\theta_{A}}\overrightarrow{j}
\end{equation}

The \(\overrightarrow{AM}\) vector is calculated through Equation \eqref{AM}, where \(A\) represents the location of the DF spring attachment point connection to toe joint of the prosthesis, $l_{m}$ is the length of the \(\overrightarrow{AM}\), and $\theta_{in}$ is the initial angle between \(\overrightarrow{AM}\) and  \(\overrightarrow{AB}\).

\begin{equation}{\label{AM}}
	\overrightarrow{AM}=-l_{m}\cos{(\theta_{in}-\theta_{A})}\overrightarrow{i}+l_{m}\sin{(\theta_{in}-\theta_{A})}\overrightarrow{j}
\end{equation}

The \(\overrightarrow{MO}\) vector is calculated using Equation \ref{MO}. In this equation \(O\) represents the origin point of the circular shape of the slider path.

\begin{equation}{\label{MO}}
	\overrightarrow{MO}=x_{MO}\overrightarrow{i}+y_{MO}\overrightarrow{j}
\end{equation}

The \(\overrightarrow{OB}\) vector is calculated using Equation \eqref{OB}, where \(B\) represents spring attachment point, and $\theta_{r}$ is the rotation \(\overrightarrow{OB}\) towards sliding path.

\begin{equation}{\label{OB}}
	\overrightarrow{OB}=-r\cos{\theta_{r}}\overrightarrow{i}+r\sin{\theta_{r}}\overrightarrow{j}
\end{equation}

The \(\overrightarrow{BP}\) vector is calculated using Equation \eqref{BP}, where \(P\) represents the position of \(B\)'s rotation position.

\begin{equation}{\label{BP}}
	\overrightarrow{BP}=-y_{B}\sin{\theta_{A}}\overrightarrow{i}+y_{B}\cos{(1-\theta_{A})}\overrightarrow{j}
\end{equation}

The \(\overrightarrow{AP}\) vector which is the summation of the vectors from  \(\overrightarrow{FM}\) to \(\overrightarrow{BP}\) represents the position vector of the DF springs. The length of this vector is determined as follows,

\begin{equation}{\label{AP}}
	{\left |\overrightarrow{AP} \right|}= {\left| \overrightarrow{AM}+\overrightarrow{MO}+\overrightarrow{OB}+\overrightarrow{BP} \right |}
\end{equation}
 
To determine the torque vector in the ankle joint, it is needed to do the cross product between the DF springs position vector and the force vector.

The force vector of the DF mechanism is determined through the stiffness of the DF springs and the deflection in these springs. The deflection in the DF springs ($\Delta x$) is determined as follows,

\begin{equation}
	\overrightarrow{\Delta x} = \overrightarrow{AP} - \overrightarrow{AB}
	\label{deflection}
\end{equation}

Also, the force vector generated by the DF springs is determined as follows,

\begin{equation}
	\mathbf{F} = k\overrightarrow{\Delta x}
	\label{force_calc}
\end{equation}

The generated torque in the ankle joint due to the DF mechanism ($\tau_a$) is determined as follows, 

\begin{equation}
	\overrightarrow{\tau_a} = \overrightarrow{FP} \times \mathbf{F}
	\label{tau_A}
\end{equation}

To provide a human-like power profile and ensure sufficient energy for the push-off phase, the remaining energy in the ankle joint is intended to be supplied by an additional EES mechanism. In the EES mechanism, a spring is compressed using a ball-screw motor during the whole gait cycle except push-off phase and release the stored energy to the ankle joint at this phase to ensure the torque and power requirements of the ankle joint in a cyclic manner throughout the walking.

In the EES mechanism, a ball-screw actuator compresses the elastic element and stores additional energy during the gait cycle, except during the push-off phase, which represents about 20\% of the cycle. In this way, the high power demand of push-off is distributed over a longer period, allowing the use of a low-power motor. During compression, the spring is aligned with the ankle axis so that it does not apply torque to the joint. At push-off, the trajectory of the spring is shifted, enabling the stored energy to generate torque about the ankle joint. Similar to the DF mechanism, the EES mechanism incorporates an arc-shaped sliding path. This arc is designed so that the spring length remains constant at maximum dorsiflexion, ensuring smooth operation. At the start of push-off, when the ankle reaches maximum dorsiflexion, the spring attachment point moves along the arc to its final position, releasing the stored energy as torque. The relocation of the spring attachment is achieved with minimal energy loss by adjusting the arc’s center using a dedicated adjustment part. This component also allows the orientation of the sliding path to be tuned according to an individual’s maximum dorsiflexion angle.

\begin{figure}[htbp]
	\centering
	\includegraphics[height=50mm]{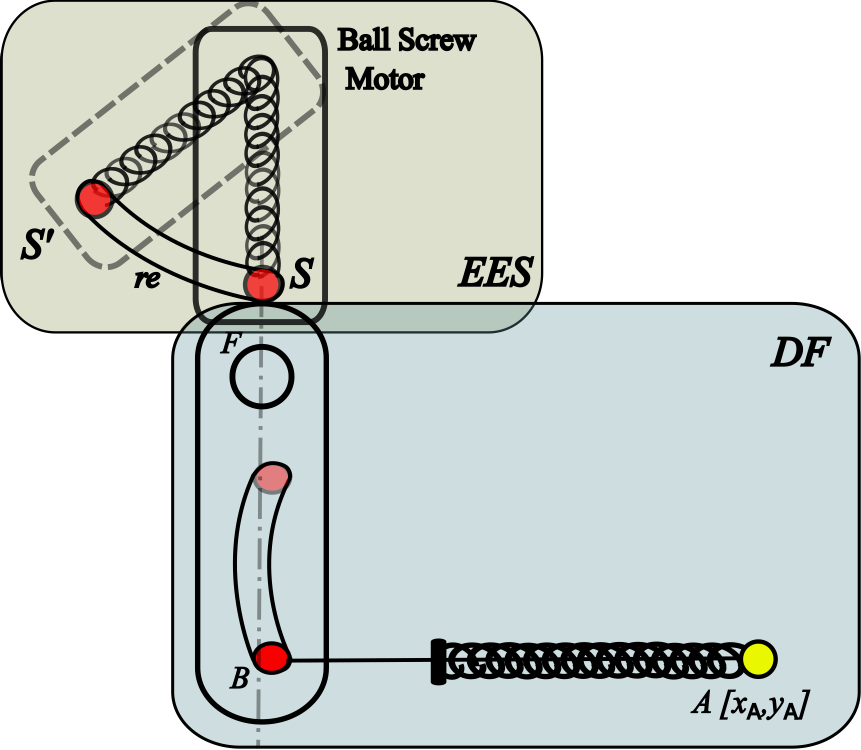}
	\caption{The schematic model of the EES mechanism}
	\label{EES_mechanism}
\end{figure}

The schematic model of the EES mechanism is shown in Fig. \ref{EES_mechanism}. In this figure, the $r_{e}$ represents the orbital radius of the EES mechanism, and $S$ and $S^{'}$ points represent the initial and final attachment locations of the EES spring. 

The $\overrightarrow{SS'}$ vector represents the moment arm of the EES mechanism and is determined as follows,

\begin{equation}{\label{SS'}}
	{\overrightarrow{SS'}}= -r\sin{\theta_{r}}\overrightarrow{i}+r(1-\cos{\theta_{r}})\overrightarrow{j}
\end{equation}

In order to determine the generated torque by the EES mechanism, it is required to calculate the generated force vector by this mechanism, which is determined as follows,

\begin{equation}{\label{FVec}}
	\overrightarrow{FE} = \mid F_{ES} \mid \sin{\theta_{r}}\overrightarrow{i} +  \mid F_{ES} \mid \cos{\theta_{r}}\overrightarrow{j}  
\end{equation}

where $\theta_{r}$ is the angle of the spring attachment point rotation towards sliding path arc. The \(F_{ES}\) is the magnitude of the force vector and is determined as follows,

\begin{equation}{\label{F_ES}}
	|F_{ES}|=K_{ES}\Delta x
\end{equation}
In this equation \(K_{ES}\) is the stiffness for the spring of the EES mechanism and the $\Delta x$ is deflection of this spring, which is compressed by the motor.

Finally, the generated torque around the ankle joint by the EES mechanism isdetermined as follows,

\begin{equation}{\label{tau_es}}
	{\overrightarrow{\tau_{ES}}}={\overrightarrow{SS'}}\times{\overrightarrow{F_{ES}}}
\end{equation}

The generated total torque around the knee joint by the RoboANKLE prosthesis is the summation of the torques generated by the DF and EES mechanisms. The simulated total quasi-static torque value around the ankle joint, is shown in Fig. \ref{subfig:df_ees_torque}. In this figure the generated torque by the DF mechanism is shown in blue and the total generated torque is shown in black. These values are compared to the natural ankle fast walking torque which is shown green.

The power generation of the DF and EES mechanisms compared to the natural ankle fast walking power is illustrated in Fig. \ref{subfig:df_ees_power_generation}. In this figure, natural ankle power flow is shown in green, and the generated power by the DF and EES mechanisms is shown in blue and red, respectively. Also, the total power around the ankle joint is shown in black. The RMSE  value between the natural ankle power and the power generated by the prosthesis is 12\%. The simulation with the addition of the proposed mechanisms resulted in a power profile closely matching the natural ankle power profile, with an achieved 97.8\% correlation.

\begin{figure}[h]
	\centering
	\includegraphics[height=60mm]{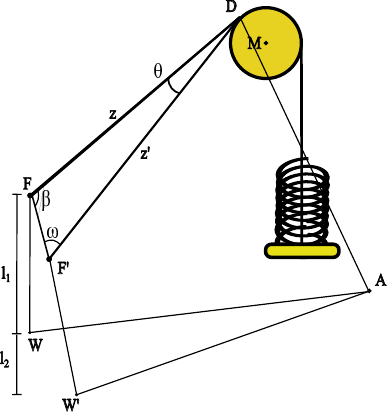}
	\caption{The schematic model of the reset spring}
	\label{Reset_Spring}
\end{figure}

The resetting spring in the ankle joint would allow the foot to return to its neutral position after the maximum push-off occurs. After push-off, in the late swing phase, the compressed reset spring returns the ankle to its neutral state (0\textdegree) that provides toe clearance. The spring under consideration is determined by comparing the potential energy differential between the prosthesis in its neutral position and complete push-off. Potential energy difference used for the determining the spring stiffness between the maximum plantar-flexion and the neutral angle of the ankle, which generates the necessary torque to elevate the RoboANKLE. The schematic model of the rest spring is shown in Fig. \ref{Reset_Spring}. In this figure, the $F$ is the spring attachment point in the neutral state of the ankle, and $F'$ is the final state of the ankle where maximum push-off in the push-off phase. The required spring stiffness and the generated torque are determined through Equations \ref{DF} - \ref{taur}. The $\overrightarrow{DF}$ is the force vector of the the reset spring, from the potential energy difference the required force is found.

\begin{figure}[htbp]
	\begin{center}
		\includegraphics[width=0.3\linewidth]{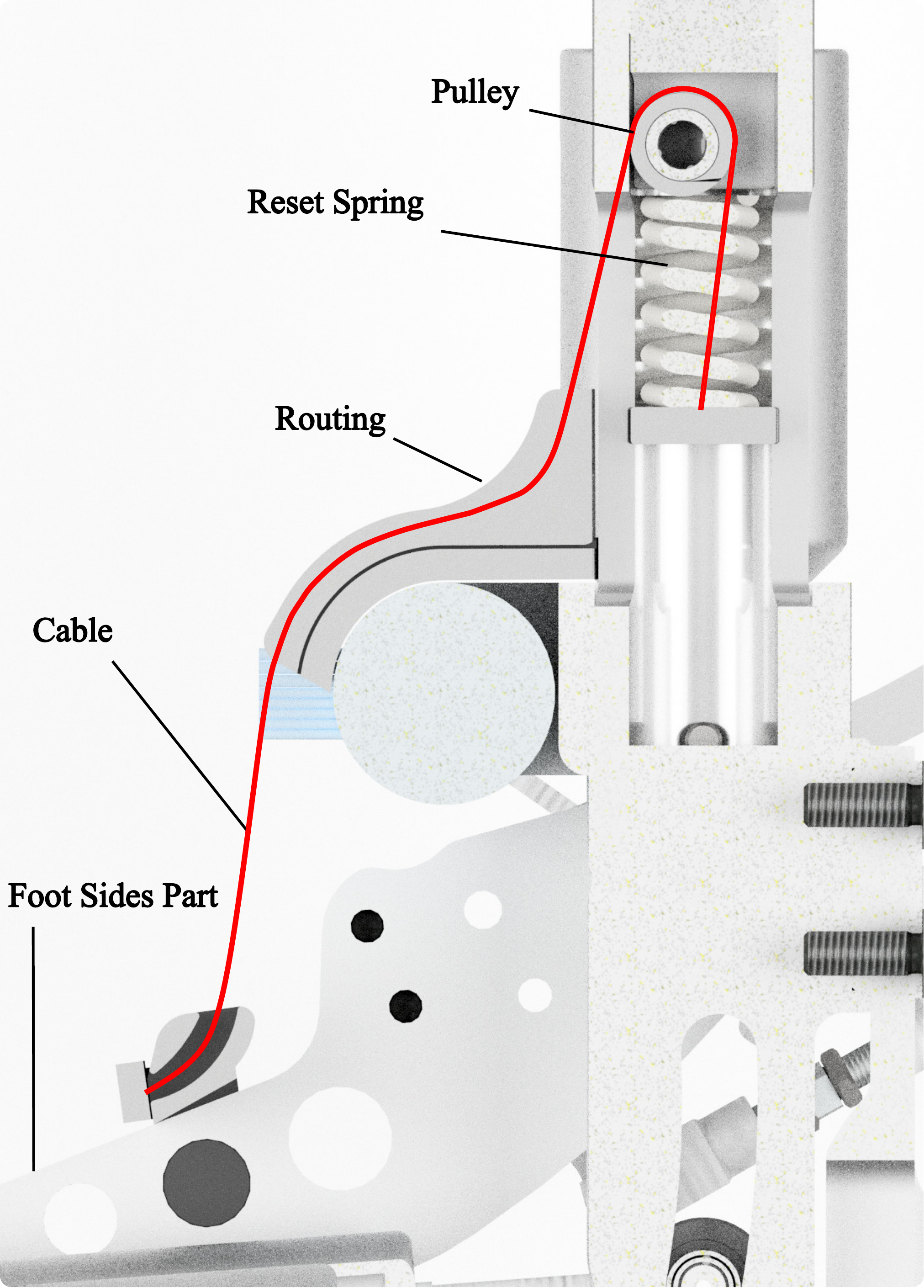}
		\caption{Reset spring mechanism in RoboANKLE}
		\label{reset_springmech}
	\end{center}
\end{figure}

The reset spring mechanism is presented in Fig. \ref{reset_springmech}. In the figure, red line cable that compresses the "Reset Spring" when "Foot Sides Part" makes plantar-flexion is shown in red, this is also means RoboANKLE is making the plantar-flexion. At first, steel material is selected as the cable material, but due to its rigidity, it is replaced with a Steelex fishing line which can carry up to 20 kg. The cable tip at the spring side compresses the spring with a cap, this generated force is rotated around the pulley. The cable passes through routing part, and connects to the foot sides part. This way, in the plantar-flexion motion of the RoboANKLE, the mechanism rotates to its neutral position.

The $\mathbf{F}$ is calculated using Equation \ref{DF}, where $\mathbf{F}$ is the force vector, $m$ is the mass of the foot compartment of the RoboANKLE, $g$ is the gravity acceleration, and $l_{2}$ is the height difference.

\begin{equation}\label{DF}
	\mathbf{F} = \dfrac{mg(l_{2})}{ \mid FF' \mid \cos{\beta}} 	
\end{equation}

The $\Delta x$ is calculated using \ref{deltax}, where $\Delta x$ is the spring displacement.  

\begin{equation}\label{deltax}
	\overrightarrow{\Delta x} = \overrightarrow{DF'} - \overrightarrow{DF}
\end{equation}

The $k$ is calculated using \ref{springcoef}, where $k$ is spring stiffness.

\begin{equation}\label{springcoef}
	k = \frac{F }{\Delta x}
\end{equation}

The $\overrightarrow{AF'}$ is calculated using \ref{AF'}.

\begin{equation}\label{AF'}
	\overrightarrow{AF'} = \mid \overrightarrow{AF} \mid \cos(\theta)\overrightarrow{i} + \mid \overrightarrow{AF} \mid \sin(\theta)\overrightarrow{j}
\end{equation}

The $\overrightarrow{DF'}$ is calculated using Equation \ref{DF'}.

\begin{equation}\label{DF'}
	\overrightarrow{DF'} = \overrightarrow{DA} - \overrightarrow{AF'}
\end{equation}

The $\mathbf{F}$ is calculated using \ref{matbfF}. 

\begin{equation}\label{matbfF}
	\mathbf{F} = |\mathbf{F}| \cdot \frac{\overrightarrow{DF}}{|\overrightarrow{DF}|}
\end{equation}

The $\tau_{r}$ is calculated using Equation \ref{taur}, where $\tau_{r}$ stands for torque generated by the reset spring. 

\begin{equation}\label{taur}
	\overrightarrow{\tau_{r}} = \overrightarrow{AD} \times \mathbf{F}
\end{equation}

\begin{figure}[h]
	\centering
	\subfloat[\scriptsize]{
		\includegraphics[width=0.56\textwidth]{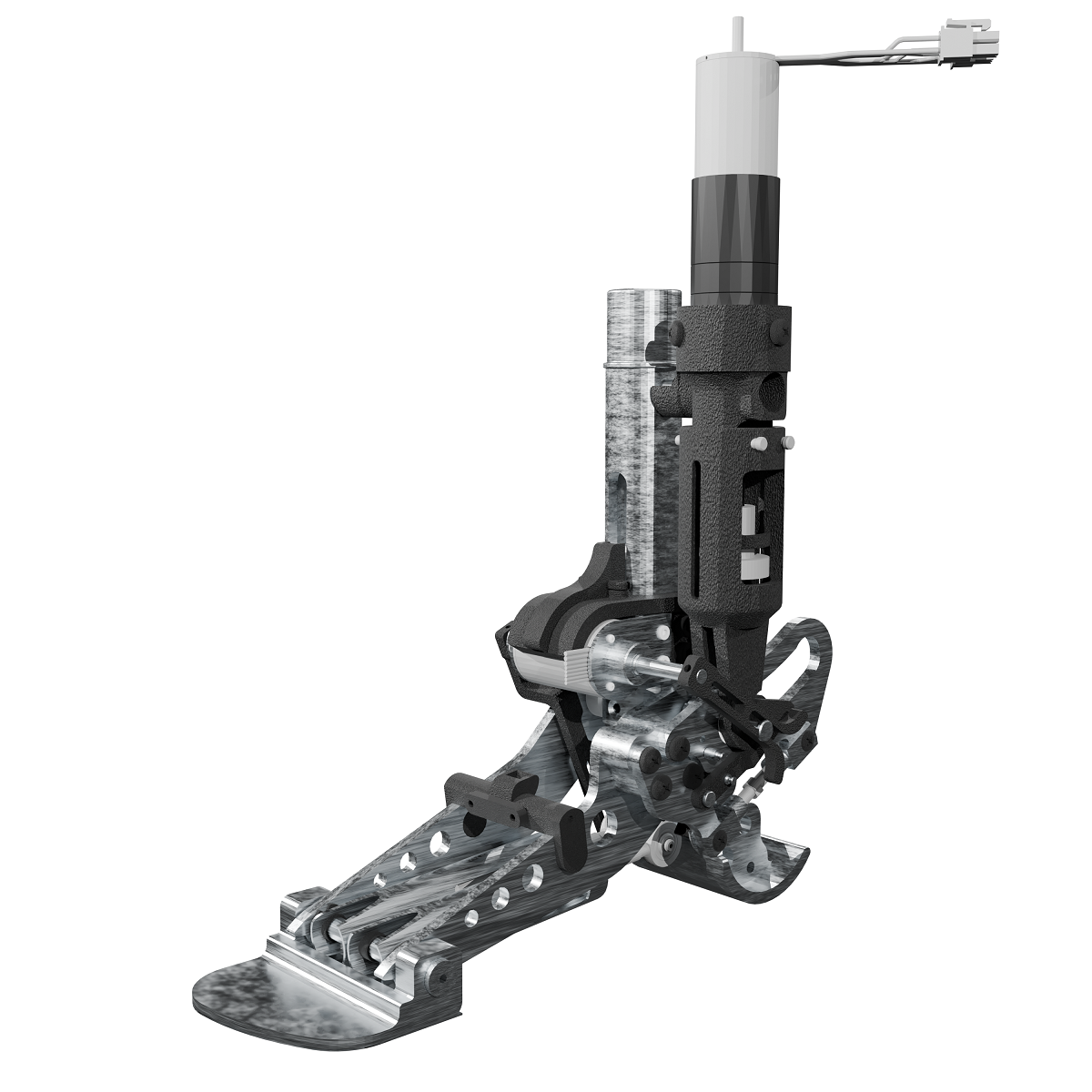}
		\label{subfig:cadmodel}
	}
	\subfloat[\scriptsize]{
		\includegraphics[width=0.34\textwidth]{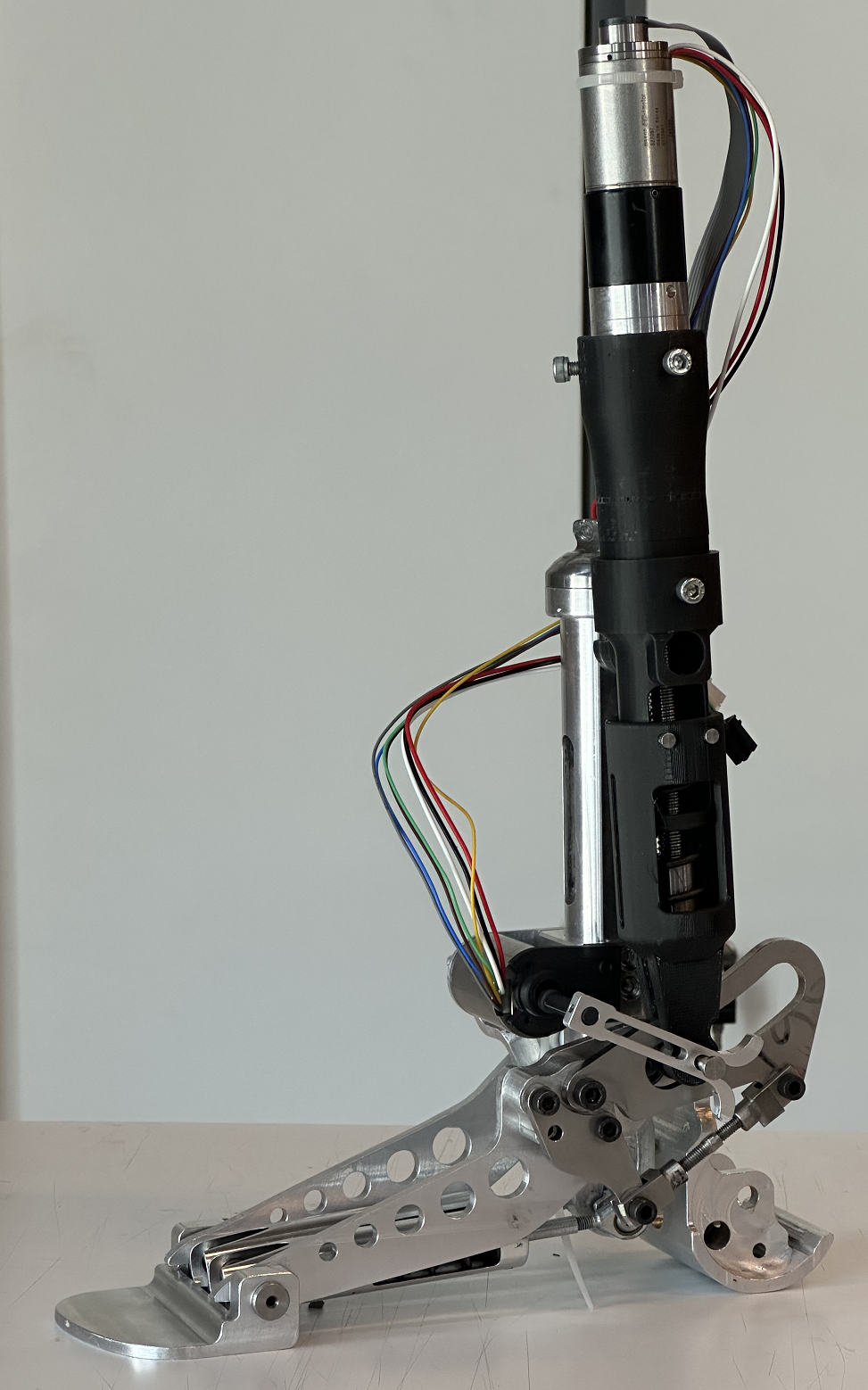} 
		\label{subfig:prototype}
	}
	\caption{The CAD model of the roboANKLE prosthesis (a), and the manufactured version of the RoboANKLE (b)}
	\label{Prototype}
\end{figure}

The CAD model of the RoboANKLE is shown in Fig.\ref{subfig:cadmodel}. In this figure, all the implemented mechanisms, such as the DF and EES mechanisms, are shown.

\subsection{Kinematic and Kinetic Analyses}

In order to evaluate the dynamic behavior of the RoboANKLE during the gait cycle, its dynamic simulations are conducted through the MSC Adams (MSC Software Corporation, USA) environment. Inertial parameters are extracted from the CAD model of the RoboANKLE. In modeling, natural ankle angles are incorporated into the system using AKIMA interpolation, which allows the model to replicate the ankle motions. In the simulation, the EES system spring is placed in compressed state, as the simulation reaches the push-off phase, the spring attachment slides to the predefined position. For the DF spring, a custom spring function is implemented, which ensures the application of the spring force in the mechanism when it is compressed.

The change in torque profile of the RoboANKLE prosthesis at the ankle joint during the walking cycle in comparison to the natural ankle torque profie is shown in Fig. \ref{subfig:adams_torque_power} (top). In this figure, the green line represents the torque of the natural ankle, while the purple line represents the ADAMS simulation results for the torque profile of the RoboANKLE. 
Figure \ref{subfig:adams_torque_power} (bottom) compares the power flow at the ankle joint during walking between the natural ankle and the RoboANKLE. The green line represents the natural ankle power flow, while the purple line represents the power flow of the RoboANKLE.

 \section{Development of the RoboANKLE}

\subsection{Instrumentation}

In this part of the study, the selection of the critical elements for the RoboANKLE's instrumentation is discussed. These elements are actuator system, Real-Time (RT) microprocessor with analog and digital converters, and battery selection.\\
In order to determine the required energy by the EES mechanism and its actuator system, the energy difference between the natural ankle and the DF spring mechanism generated energy during the push-off phase of the gait cycle is considered. This phase is characterized by a high energy requirement to move the body forward. Applying the remaining portion of the energy, will deliver full support of the RoboANKLE prosthesis during the push-off phase. The actuator system of the EES system includes a Maxon EC-i30 motor integrated with a GP32S ball screw gear, featuring a 4.8:1 reduction rate. To have a lightweight actuator system the ESCON Module 50/5 is selected as the motor driver for this mechanism which has a weight of 12 g. Additionally, the ENC 16 easy 512 cpt incremental encoder is selected for measuring velocity of the motor. 

Raspberry Pi 3 Model B (RPi) is selected as the microprocessor for the actuator system. The choice is motivated by its low cost, compact size, and ease of development, as well as its ability to support real-time control tasks through its quad-core CPU and extensive GPIO interface. These features make it a suitable platform for rapid prototyping of robotic systems while ensuring sufficient computational power for sensor integration and control algorithms. Since the RPi does not support analog signal input/output, a high precision Analog-to-Digital/Digital-to-Analog (AD/DA) (Analog-to-Digital/Digital-to-Analog) (Waveshare) converter is integrated to provide the required analog signals. To ensure appropriate behavior during critical gait events such as heel strike and push-off, a sampling rate of 1~kHz is adopted for real-time control of the prosthesis. The control algorithm of RoboANKLE regulates the EES mechanism by processing sensor data, generating control decisions, and commanding the robotic components to respond appropriately during gait.

To be able to mobile use of the RoboANKLE and without a fixed power source, a battery is required. Also, in order to meet all task requirements of the user throughout the day, the selected battery must have the appropriate capacity. The required energy for each step can be determined as the multiplication of the daily step and work done in each step. Then the required capacity of the battery is determined as follows,
\begin{equation}
	A_h= \frac{E_{total_{Wh}}}{V}
\end{equation}
Where, $E_{total_{Wh}}$ is the total required energy in terms of Watt-hour, and V is the nominal voltage of the selected motor.
Considering the energy required for each walking cycle at ankle joints (10 J/step), the total energy required for 5000 walking cycles is 50000 J. Since the nominal voltage of motors in the RoboANKLE prosthesis is 24 V, one of the best-performing batteries on the market, the MoliCell Lithium-ion rechargeable battery (3.7 V nominal and a maximum voltage of 4.2 V), is selected. Each battery has a capacity of 4200 mAh (362880 J), and when 8 lithium-ion batteries are connected in series, they provide a maximum of 33.6 V and a continuous current of 10 A. Thanks to the selected battery, the aforementioned daily activities can be carried out with a single charge. The charge rate of the batteries starts at 4.2 V and drops to 2.5 V at the end of discharge. The mass of each battery cell is 48 g, and the total mass of the battery pack is 414 g.

\subsection{Structural Analysis and Prototyping}

\begin{figure}[h]
	\centering
	\hspace{-3mm}
	\subfloat[]{
		\includegraphics[width=0.45\textwidth]{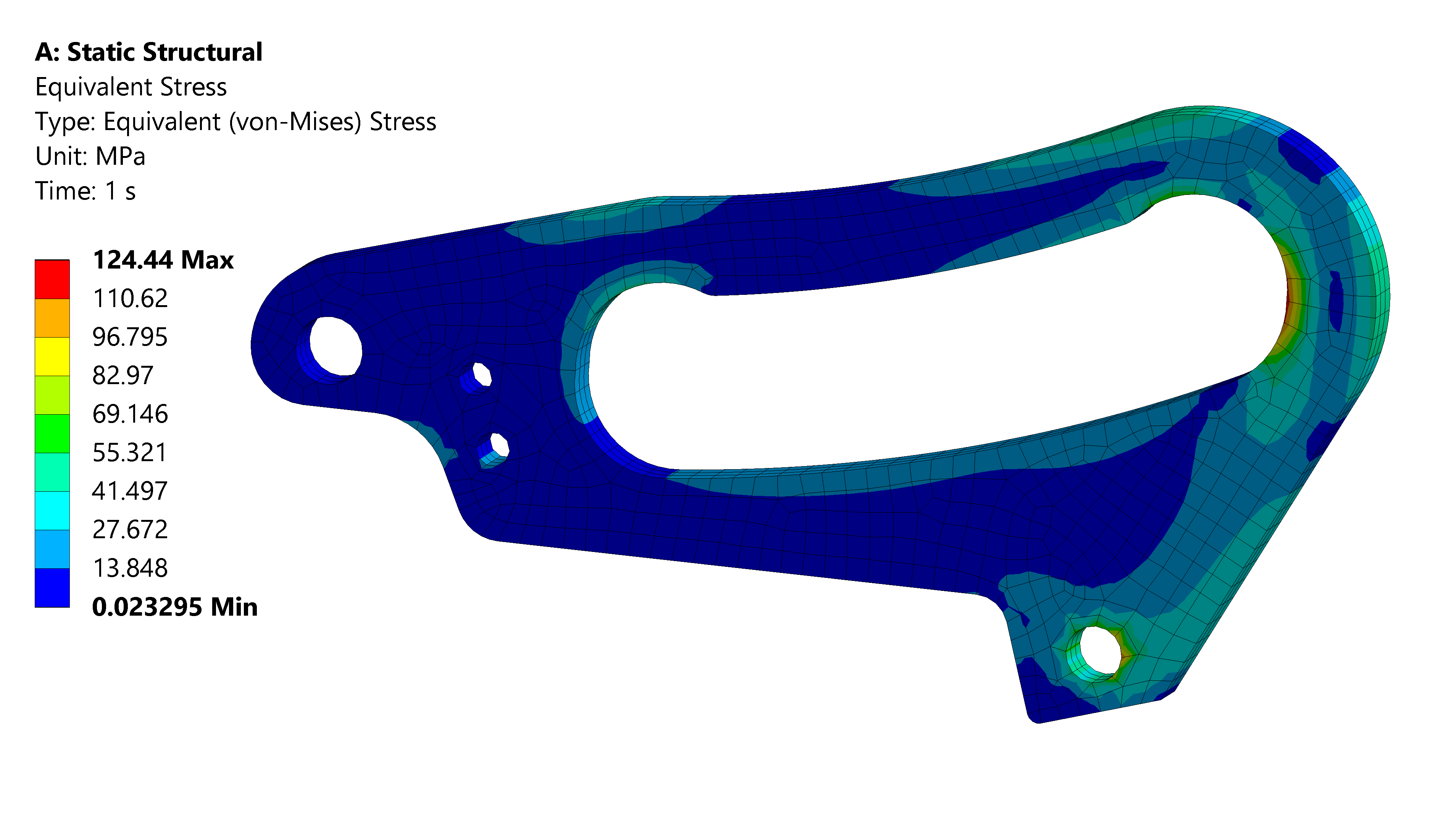}
		\label{slidinganalysis}
	}
	\hspace{-3mm} 
	\subfloat[\scriptsize]{
		\includegraphics[width=0.45\textwidth]{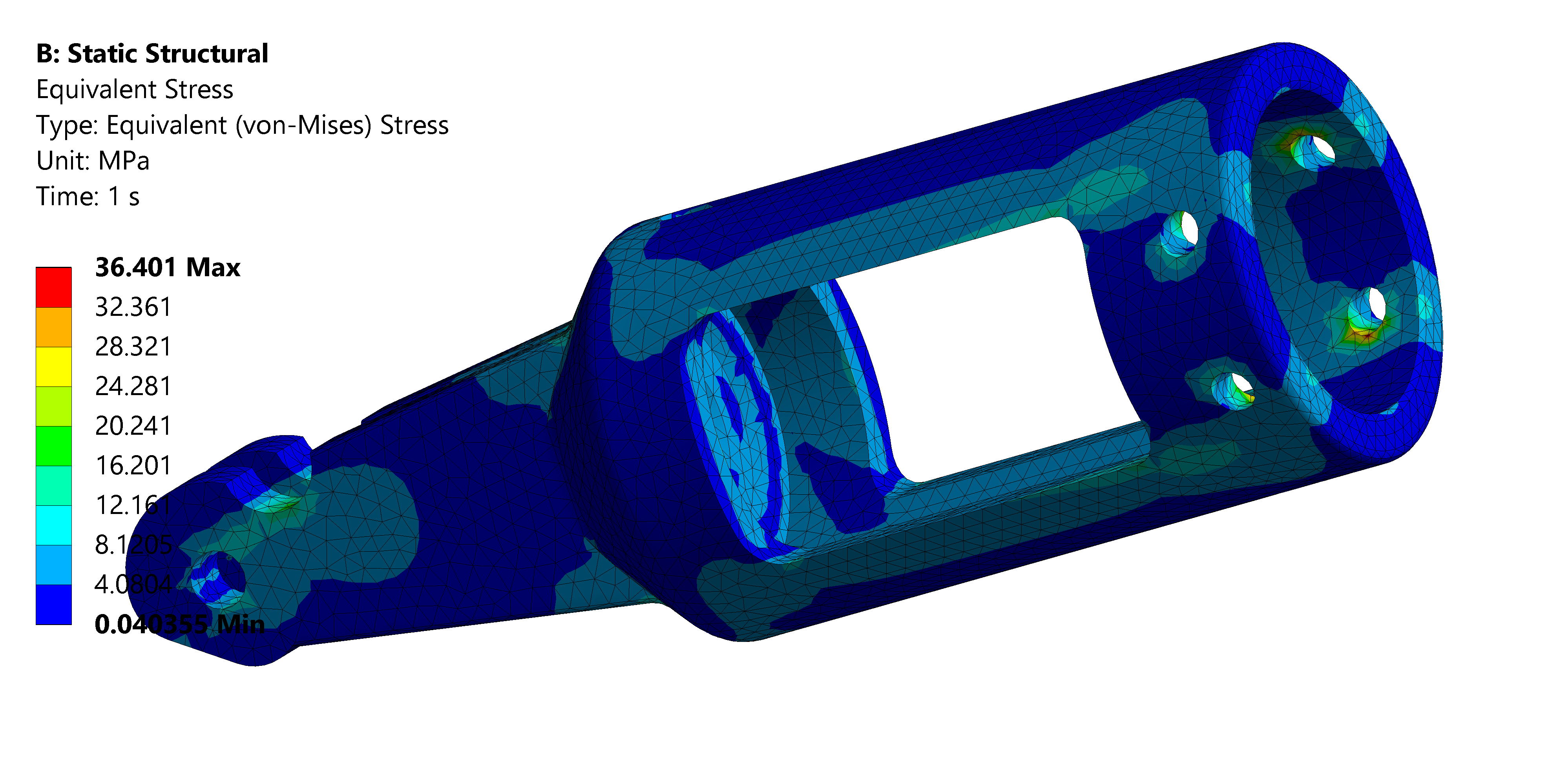}
		\label{eesanalysis}
	}

	\caption{The stress analysis result for the the slider part (a) and lower tube part of EES mechanism (b).}
	\label{Structural-Analyses-EES}
\end{figure}

\begin{figure}[h]
	\centering
	
	\includegraphics[width=\textwidth]{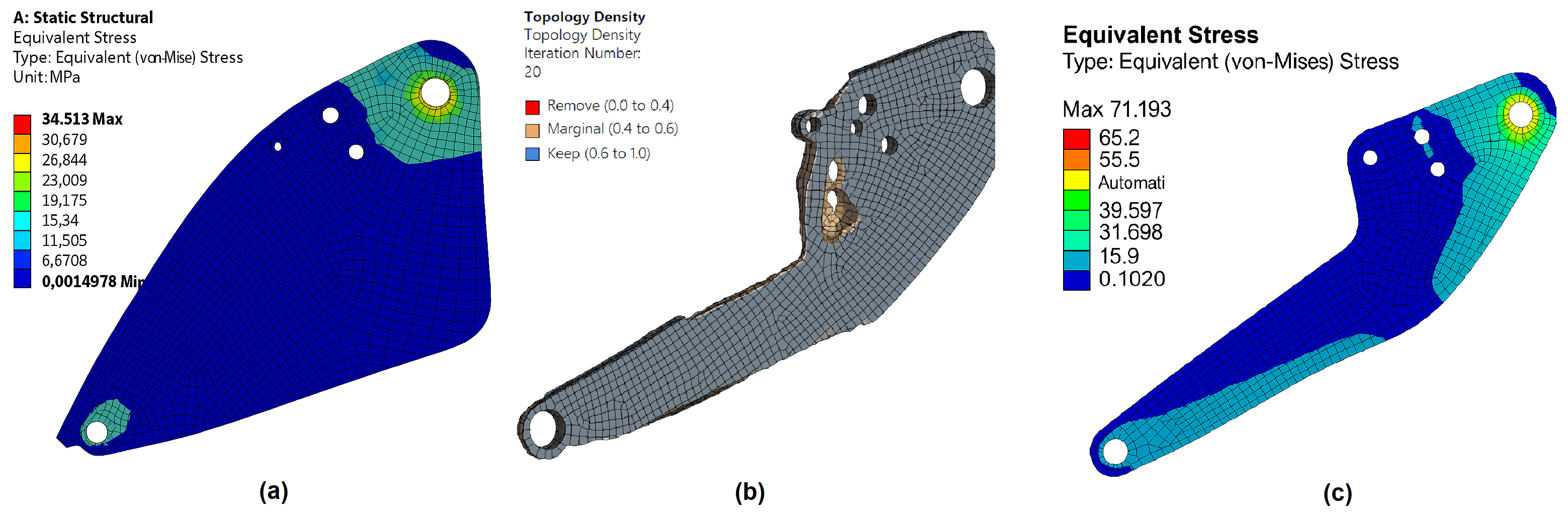}
	
	\caption{The structural analysis and the topology optimization of the RoboANKLE foot side part, the initial part (a), the suggested topology optimization for this part (b), the final optimized foot side part (c).}
	\label{Foot_Sides}
\end{figure}

Weight plays a critical role in the design of powered prostheses, influencing both the mechanical performance and user comfort.  The prosthesis has weight-bearing parts to counteract the high torque generated during walking. Natural ankle joint torque values reaches 130 N$\cdot$m during fast walking (1.6 m/s). The forces acting on different parts of the RoboANKLE  are derived from dynamic simulations in ADAMS. Structural analysis and topology optimization conducted of the RoboANKLE parts to ensure strength of the RoboANKLE while providing a lightweight system through ANSYS (ANSYS, USA). The slider part  of the EES mechanism is one of the load-bearing parts in this mechanism. Total of 900 N force is exerted on this component by the motorized spring. The structural analysis of this component is shown in Fig. \ref{slidinganalysis}. The maximum Von-Mises stress value for this part is 124.44 MPa. Also, from the ADAMS simulation it is resulted that 900N force is exerting on the lower tube part of the EES mechanism from the motorized spring. The results of structural analysis for this is shown in Fig. \ref{eesanalysis}, where a maximum Von-Mises stress value of 36.401 MPa is seen on this component. One of the biggest parts in the prosthesis is foot side part, hence the topology optimization for this component in order to decrease its mass and consequently the final mass of the RoboANKLE is conducted using ANSYS Topology Optimization tool. In the Fig. \ref{Foot_Sides} (a) to \ref{Foot_Sides} (c), the structural analysis and the topology optimization of the RoboANKLE foot side part is shown. With the topology optimization the parts mass is decreased. While doing that safety factor is taken into consideration. In the Fig. \ref{Foot_Sides} (a), the structural analyses of the initial design of the foot side part is presented. Then, the topology optimization with mass minimization criteria is conducted as presented in Fig. \ref{Foot_Sides} (b). Finally, the output of the topology optimization is revised and the final CAD model of the foot side part and its structural analysis is shown in Fig. \ref{Foot_Sides} (c). Aluminum 7075 was chosen due to its high strength-to-weight ratio, a property that makes it widely used in aerospace and high-performance mechanical structures \cite{davis1993aluminum}. According to the Structural analysis results, certain components of the EES mechanism are chosen to be manufactured from carbon fiber-supported Onyx material. The Onyx parts are manufactured with a Markforged X7 (Markforged, USA) 3D composite printer. The amount of carbon fiber layers are implemented over the results indicated that applying a single carbon fiber layer to the part increased the tensile stress by up to 60 MPa \cite{Emre2023}. Considering this fact, at least 3 layers of carbon fiber are implemented in the direction of the applied force. Additionally, high-stress-bearing components materials are selected to be manufactured from AISI 302 Steel, which includes parts such as pins and shafts of the RoboANKLE.

\section{System Characterization}

In this section, the performance and characterization of the RoboANKLE are presented. Also, the control of the EES mechanism is presented. To validate the torque generation capacity of the RoboANKLE and ensure that the prototype meets the required performance fro the specific tasks, the system identification process determining the transfer functions of the actuators and the characterization of the RoboANKLE is conducted. Detailed tests and simulations are performed to analyse the behavior of the system under various conditions, focusing on the key components such as the ball screw and replacer motors. The results from these experiments are critical in assessing the overall functionality and efficiency of the prosthesis.

\subsection{System Identification}

System Identification (SI) is applied to determine the transfer function of the actuators to be implemented in their control simulations. The input signal is considered as a chirp signal that oscillates between 0.45V and 4.54V with an increasing frequency over time.
The parameters and function used to generate the chirp signals is determined as follows, 

\begin{equation}\label{time_sp}
	\begin{aligned}
		time: 0 \leq t \leq M \\
		frequency : \omega_{1} \leq \omega \leq \omega_{2} \\
	\end{aligned}
\end{equation}

\begin{equation}\label{ui}
	u(i) = A\sin{\omega_{1}t +(\omega_{2} - \omega_{1})\frac{t^2}{2M}} 
\end{equation}

Where, \(\omega_1\) and \(\omega_2\) are the start and end frequencies of the signal, respectively and \(M\) is the final period that the signal should reach at the final frequency. Additionally, Equation \eqref{ui2}, \(A\) is the amplitude of the generated signal. The instantaneous frequency which represents the frequency of signal for all the time (t) is determined as follows,

\begin{equation}\label{ui2}
	\omega_{i} = \omega_{1} +(\omega_{2} - \omega_{1})\frac{t}{M}
\end{equation}

\begin{figure}[htbp]
	\begin{center}
		\includegraphics[width=1\linewidth]{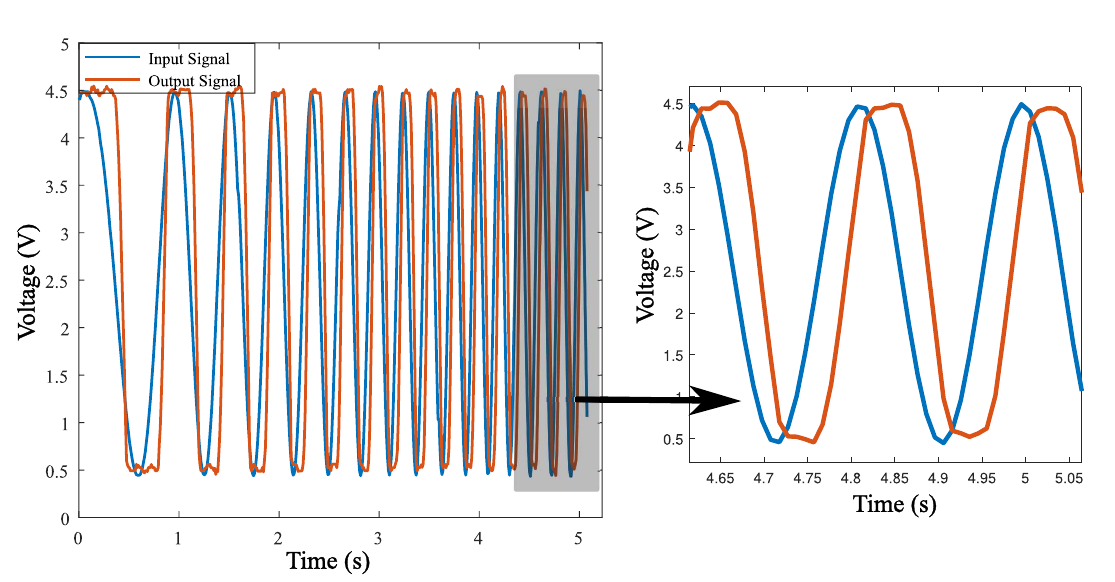}
		\caption{Input/Output voltage signals based on the chirp function for the ball-screw motor in the RoboANKLE}
		\label{chirpgraph}
	\end{center}
\end{figure}

The operating frequencies of the motors are determined based on the operating principles of the motors and the requirements of the prosthesis. While the ball screw motor is responsible for compressing the spring in the EES mechanism within period, replacer motor is responsible for rotating the moment arm around ankle joint to provide the desired torques in the system. 
In RoboANKLE, due to design limitations of the EES mechanism, the maximum deflection of the motorized spring is determined to be 20 mm. This spring is expected to be compressed within 0.9 seconds.
The working frequency calculated for the ball screw motor is 2.29 Hz, while the working frequency for the replacer motor is 16.67 Hz. To determine the SI for the motors $\omega_{1}$, $\omega_{2}$, $A$, and $M$ parameters are chosen as 0 Hz, 30 Hz, 1.35 V, and 30 s, respectively. In Fig. \ref{chirpgraph}, the input and output signals as the chirp function for the ball-screw motor is shown.

\begin{figure}[htbp]
	\begin{center}
		\includegraphics[width=0.7\linewidth]{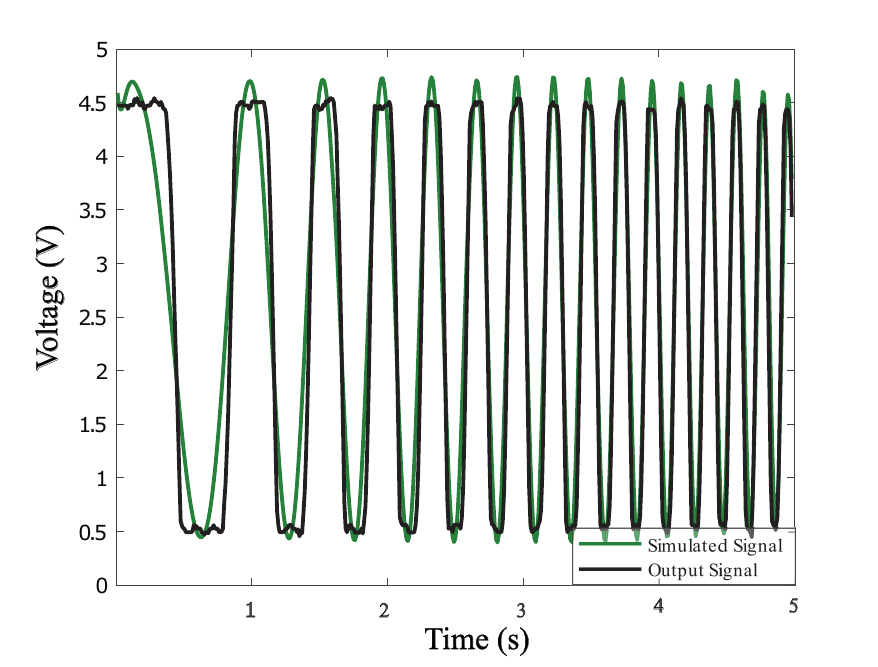}
		\caption{The performance of the ball-screw motor's transfer function}
		\label{ABS_tf_perf}
	\end{center}
\end{figure}

The performance of the defined transfer function is presented in Fig. \ref{ABS_tf_perf}. The transfer function, defined using the output signal of the ankle ball-screw motor, is compared, and the best performance was determined to be 71.24\% using Matlab's SI tool.
Using Matlab's SI toolbox, the ball screw motor's maximum frequency value was determined to be 5.26 Hz. The resulting transfer function is determined as follows,

\begin{equation}\label{tftf}
	\frac{0.1829}{s^2+0.5079s+0.1751}
\end{equation}

\begin{figure}[htbp]
	\begin{center}
		\includegraphics[width=1\linewidth]{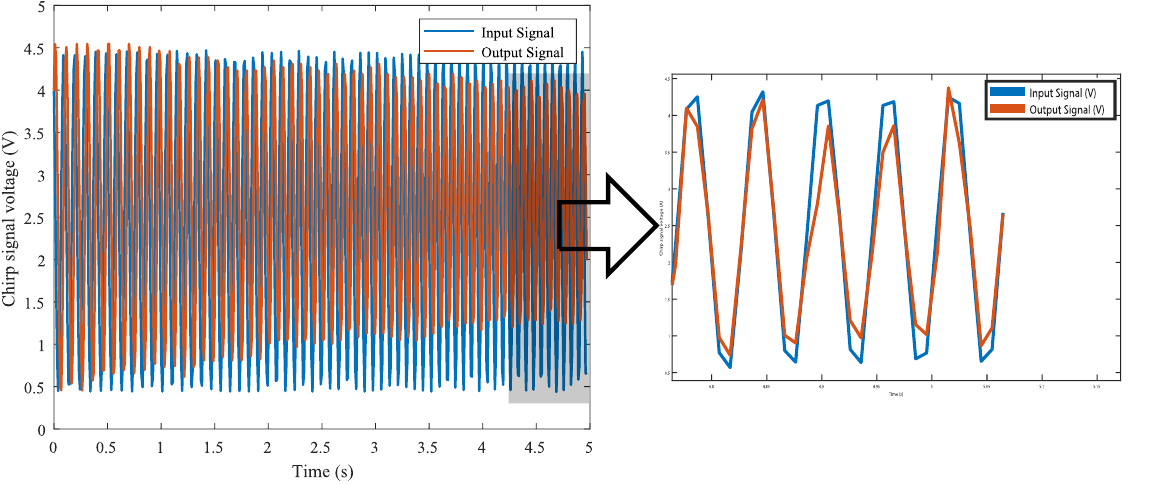}
		\caption{Input/Output voltage signals based on the chirp function for the replacer motor in the RoboANKLE}
		\label{io_replace}
	\end{center}
\end{figure}

For the replacer motor of the RoboANKLE, the input and output voltages is shown in Fig. \ref{io_replace}. In Equation \eqref{tftftf}, the transfer function of the replacer motor in the RoboANKLE is determined, based on the input and output signals. The maximum frequency value is 20 Hz, which is suitable for the motor to operate at the desired frequency of 16.67 Hz.

\begin{equation}\label{tftftf}
	\frac{0.3963}{s^2+0.9563+0.4228}
\end{equation}

\begin{figure}[htbp]
	\begin{center}
		\includegraphics[width=0.6\linewidth]{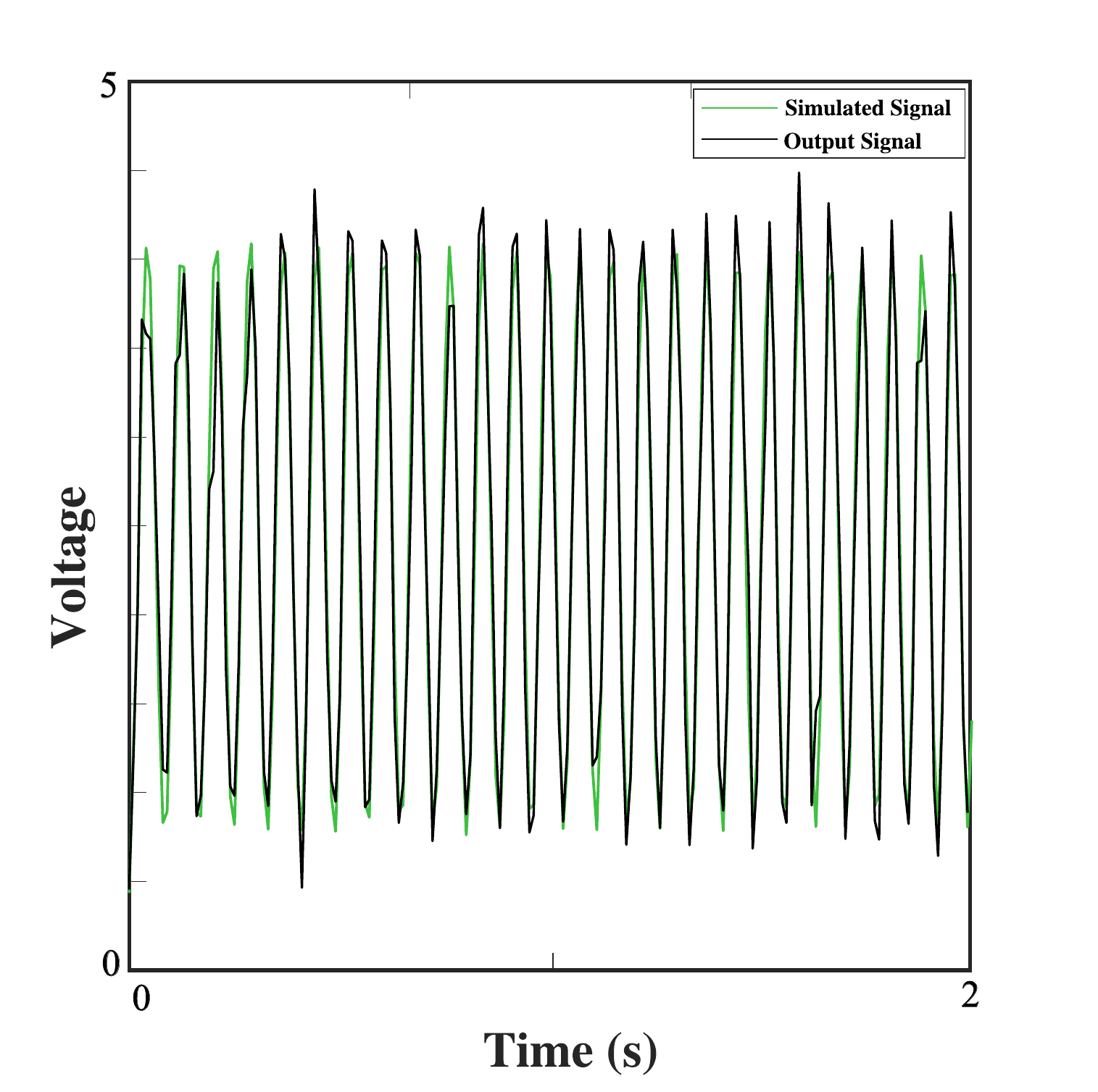}
		\caption{The performance of the  replacer motor's transfer function}
		\label{tf_perf}
	\end{center}
\end{figure}

In Fig. \ref{tf_perf}, the performance of the output signal and transfer function is compared, and this transfer functions best performance is defined as 74.26\%.

\subsection{Characterization Experiments}

In this part of the study, the characterization of the RoboANKLE to verify its capacity to generate the required torque and power during the specified tasks is discussed. To reach this goal, compression of the motorized spring experiments are conducted and capabilities of the prosthesis is evaluated. 
For the fast walking, the required current for the ball-screw motor to achieve the required torque is determined and exerted to the motor for the compressing the EES mechanism spring. The internal PID control of the motor driver is used in this phase. Then, the performance of the motor under these circumstances are measured. 

The motorized spring in the EES mechanism which has a stiffness of 45 N/mm is compressed 19 mm during this experiment. The measured value is 11\% lower than the specified value of 22 mm. This discrepancy can be attributed to the spring reaching its compression limit in physical manner, preventing further movement due to friction inside the tube. The spring compress in 0.23 seconds, which is 75\% faster than the desired value of 0.9 seconds. The compressed spring generates a force of 256.50 N. The generated torque with this spring force is approximately 38.5 N·m, which is 3.8\% less than expected. However, when additional torque is supplied with the dorsi-flexion mechanism, the system torque will be sufficient to compensate for the natural foot torque during walking. The calculated energy production is approximately 4 Joules, 12.5\% less than the desired 4.5 J. This discrepancy can be attributed to the insufficient amount of spring compression.

\begin{figure}[h]
	\begin{center}
		\includegraphics[width=1\linewidth]{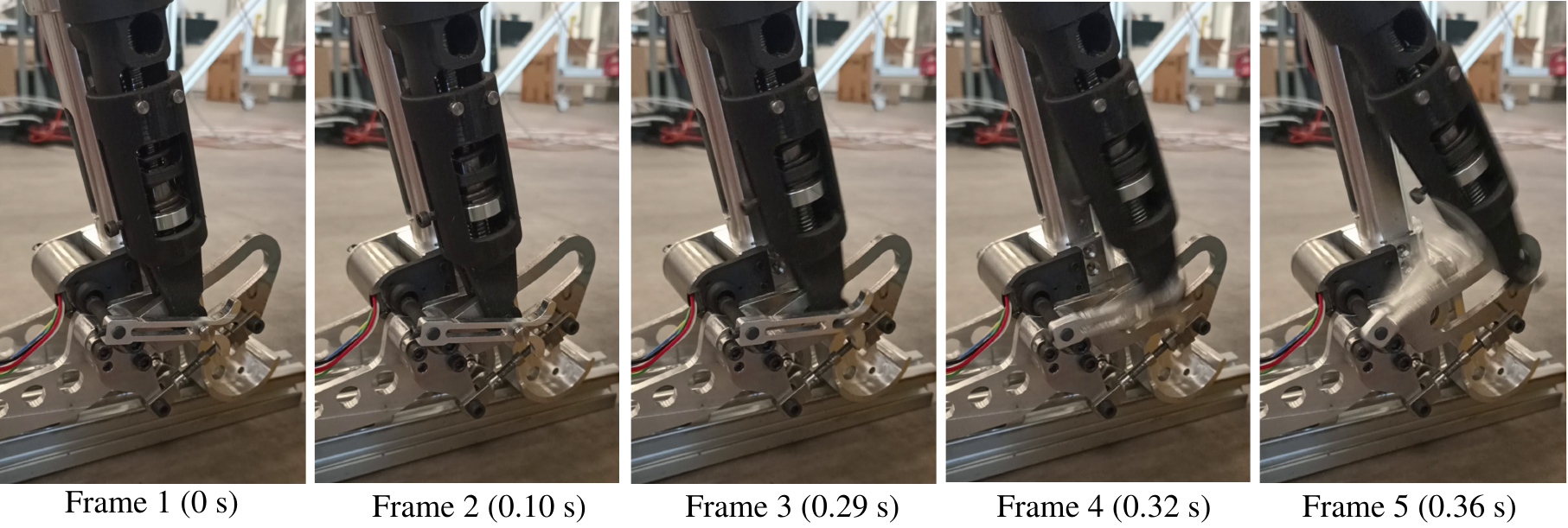}
		\caption{Characterization of the EES mechanism}
		\label{replacement_tests}
	\end{center}
\end{figure}

To evaluate the compression rate of the motorized spring in the EES mechanism, the movement of this mechanism in the specified trajectory by the displacement motor when the spring is fully compressed is evaluated. The results show that, it takes 0.23 s for the EES mechanism to fully compress the spring. This characterization is shown in Fig. \ref{replacement_tests}, and depicted from frame 1 to frame 3. Also, the displacement motor then deflected the EES mechanism in 0.13 seconds, illustrated in frames 4 to 5 of this figure.

\begin{figure}[h]
	\begin{center}
		\includegraphics[width=1\linewidth]{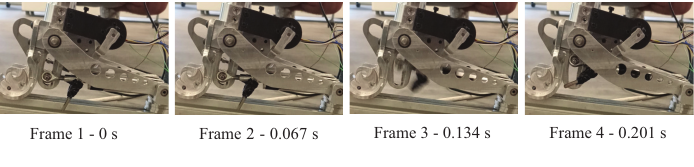}
		\caption{Characterization of the replacer motor of the DF mechanism.}
		\label{dorsimech_replacer}
	\end{center}
\end{figure}

In order to evaluate the characterization of the replacer motor in the DF mechanism, the ankle joint angle is set to the highest dorsi-flexison angle to compress the mechanism into the designated sliding slot fully. DF springs are used to manipulate the torque generated in the foot by a sliding joint attached to the foot. This sliding trajectory has a certain radius to perform this manipulation during the maximum dorsi-flexison phase in the thrust phase.
In Fig. \ref{dorsimech_replacer}, the manipulation of the fully compressed DF spring mechanism by the displacement motor is shown frame by frame. This displacement is depicted between frames 1-4, and occurs within 0.1 s. To fully compress the spring, the ankle shank is elongated with an aluminium sigma profile to 170 cm long, and the heel is fixed to the ground. This assembly is designed to apply the desired torque by bringing the DF springs to their maximum compressed state; the extended leg belt is fixed at an angle of \(9^\circ\). As a result of this dorsi-flexison of the foot joint and compression of the spring, the mechanism generates a force of approximately 2150 N. Thus, the maximum torque produced at the foot joint is 130 N\(\cdot\)mm. To evaluate the maximum total generated torque by the RoboANKLE, this amount is added the torque generated by the EES mechanism. Total torque is determined to be approximately 140 N\(\cdot\)m. The required torque in the ankle joint for a 75 kg person during fast walking is 130 N\(\cdot\)m \cite{winter}. The RoboANKLE is able to provide 7.7\% more than the natural required ankle torque in fast walking. Additionally, the displacement of the DF springs are measured as 0.13s which is less than the time required for the task (0.15 seconds). According to geometric calculations, it means that all of the energy is given before the end of the push-off phase. In the ankle, the displacement of the spring along the sliding route is aimed to be during the push-off by using the actuator. 

\subsection{Motor Control}
\begin{figure}[h]
	\begin{center}
		\includegraphics[width=0.9\linewidth]{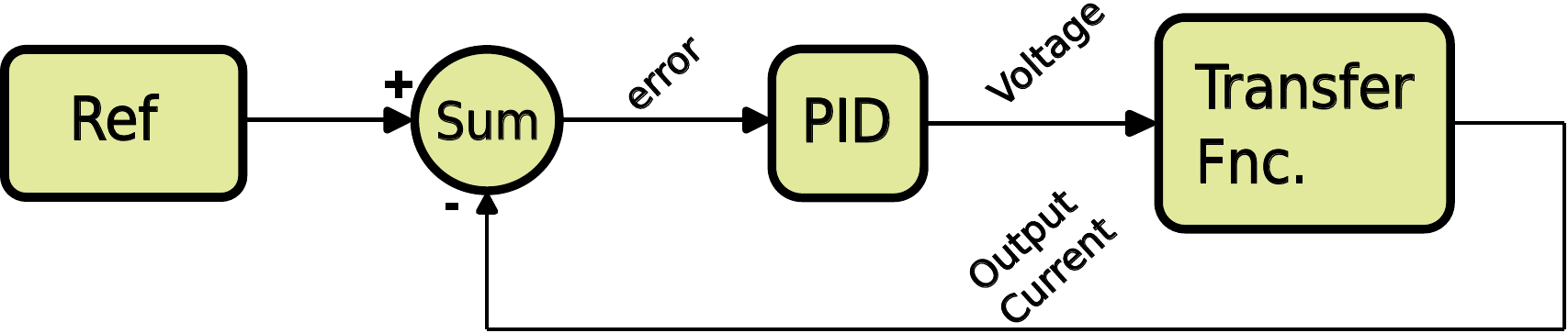}
		\caption{Control block diagram for real-time simulation.}
		\label{simu_model}
	\end{center}
\end{figure}

Block diagram to control the velocity of the actuator system is presented in the Fig. \ref{simu_model}. Using the PID control the reference and output current control of the actuator system is carried out based on the error value between these two. 

\begin{figure}[h]
	\begin{center}
		\includegraphics[width=0.8\linewidth]{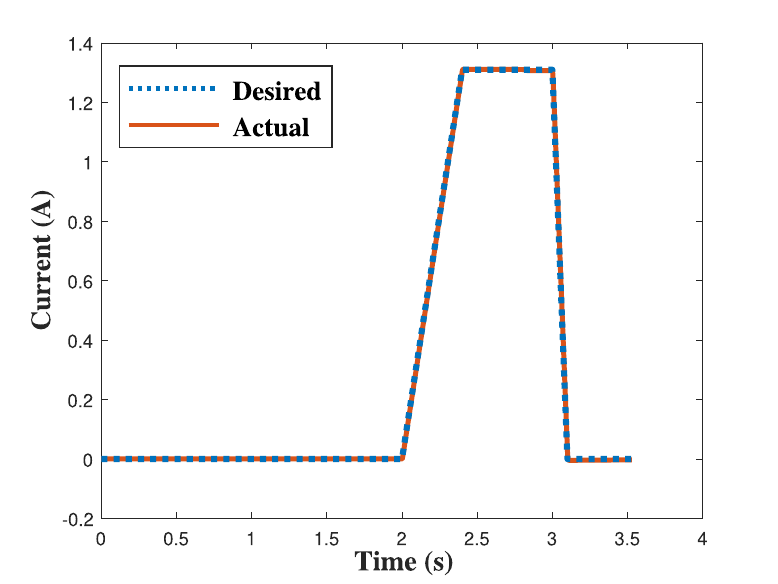}
		\caption{Results for the  real-time control simulation of the ball screw motor.}
		\label{RT-ABS}
	\end{center}
\end{figure}

Real-time simulations is performed to control current of the ball-screw motor based on the transfer functions of the actuators. The transfer function determined for ball-screw drive motor is accepted as the model for the system, and the PID method is applied to control the output torque in this model. A ramp function is applied as the input current for actuator system. The block diagram of the real-time simulation to control the output current of the actuator system is shown in Fig. \ref{RT-ABS}.

In this figure, the input and output currents of the ball-screw motor for the RoboANKLE are shown with blue and red lines, respectively. The input current is assumed to be a ramp function, starting at 2s and reaching 1.31 A in 0.4s, then dropping to zero in 0.6 s. The RMSE value between the input and output currents is determined as 6 mA. In this simulation, the P and D gain values for the PD controller are set to 1000.

\begin{figure}[htbp]
	\begin{center}
		\includegraphics[width=1\linewidth]{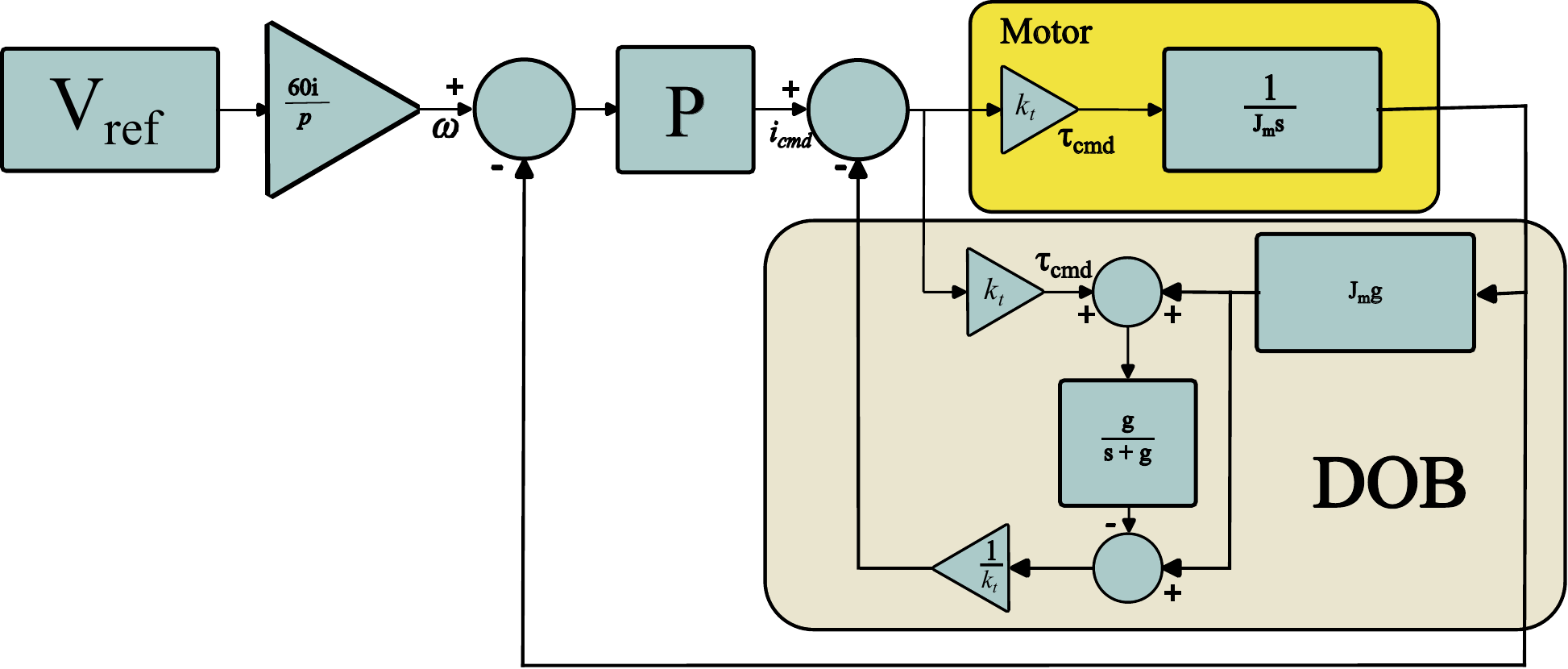}
		\caption{Block diagram of DOB velocity controller.}
		\label{DOBmodal}
	\end{center}
\end{figure}

In order to effectively compensate for external disturbances in motor speed control, enhancing the control system's robustness and stability, the DOB method is applied in combination with the PID controller. In the Fig. \ref{DOBmodal}, block diagram of the control method used is presented.

In this block diagram, \( g \) represents the cutoff frequency of the DOB's low-pass filter. The \( J_m \) is considered as a combination of the motor's gear ratio and the rotor inertia (\(j_{m}N^2\)). \( K_t \) in the block diagram converts the command current to command torque and represents the motor's torque constant. In this control method, the motor speed control is achieved by adjusting the \( g \) and \( P \) parameters. Furthermore, due to the high current demand when using a step function as the input voltage or speed in DC motors, it is replaced with a third-order polynomial function (\(velocity = at^3 + bt^2 + ct + d\)). The coefficient of this equation are determined using initial and final speed, initial and final acceleration, and initial and final noise. 

\begin{figure}[htbp]
	\begin{center}
		\includegraphics[width=0.8\linewidth]{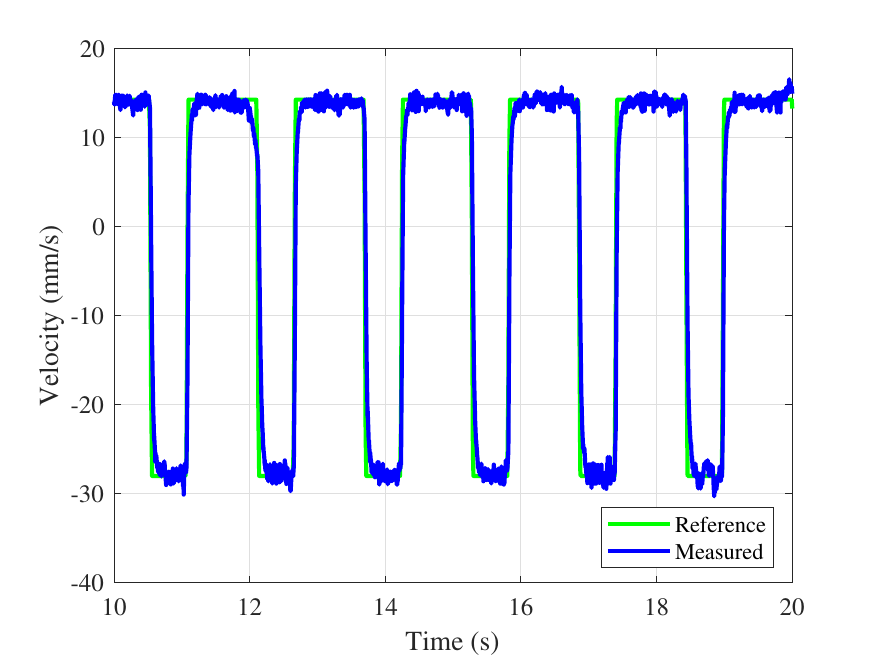}
		\caption{Velocity control results of the RoboANKLE's ball screw motor}
		\label{bsm_ankle_controlled}
	\end{center}
\end{figure}

The velocity of the ball-screw motor is desired to be controlled at 14.5 and -28 mm/s, respectively for duration of 1 s and 0.5 s. Here, the positive speed indicates compression of the spring, while negative speed indicates release of the compressed. The reference velocity and measured velocity is presented in Fig. \ref{bsm_ankle_controlled}. The green line represents the velocity reference of the ball-screw while blue line depicts velocity measurement reference. The correlation coefficient between the reference and measured values is 97.8\%.

\section{Functional Evaluations of RoboANKLE}

To evaluate the mechanical performance and the functionality of the RoboANKLE a set of experiments are conducted. In these experiments the DF mechanism to generate the torque around the ankle joint and the kinematic and dynamic functionality of the RoboANkLE during the walking are assessed.

\subsection{Mechanical Experiments}
\begin{figure}[htbp]
	\begin{center}
		\includegraphics[width=0.9\linewidth]{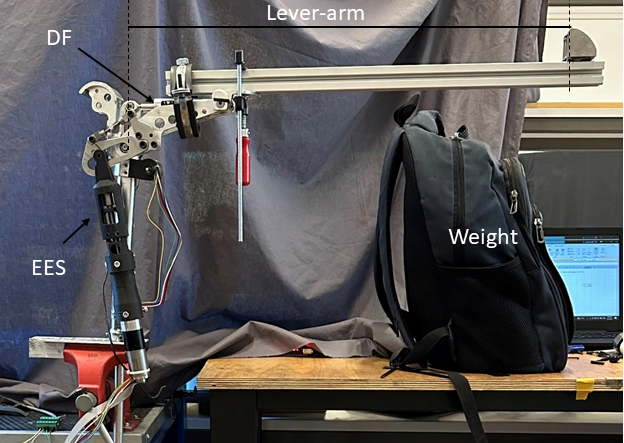}
		\caption{Test setup of the mechanical experiments.}
		\label{case_study_setup}
	\end{center}
\end{figure}

To evaluate the mechanical capacity of the DF mechanism of the RoboANKLE, a case study with this prototype is discussed in this part of the study. This experiment measures the ability of the system to provide the required maximum torque of 130 N.m in natural fast walking. To do this experiment, a setup shown in Fig. \ref{case_study_setup} is implemented. The aim of this setup is to generate torque in the ankle using weight and a lever arm to verify if the system can produce a 130 N.m torque using the DF mechanism. An aluminum sigma profile is utilized as the lever arm, adjusted to place a load distance of 60 cm. To realize the required torque in the ankle joint, a mass of 22 kg is applied within the setup. The ankle angle change is measured using potentiometer inside the hallow shaft in the experiments. 

\begin{figure}[htbp]
	\begin{center}
		\includegraphics[width=0.8\linewidth]{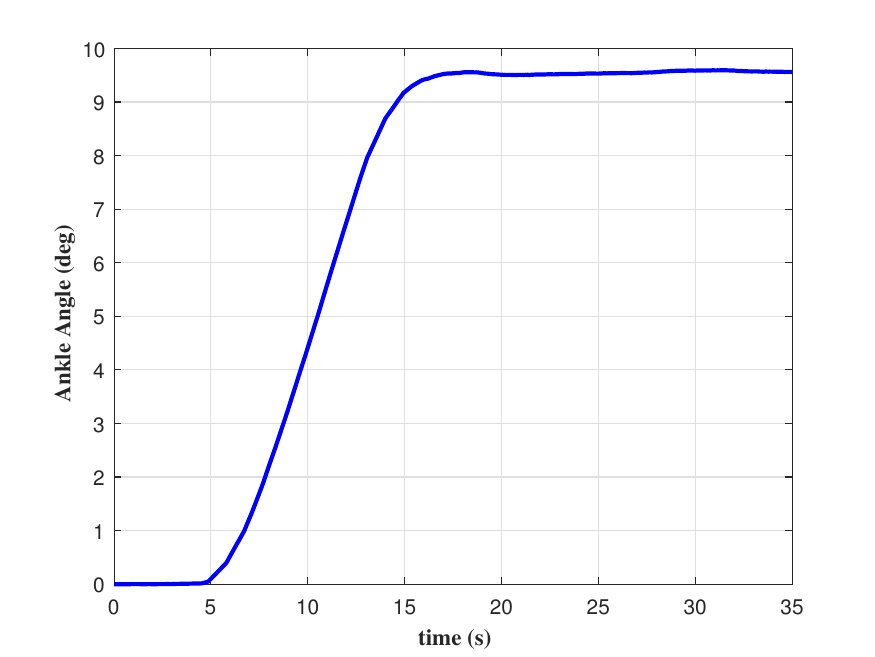}
		\caption{RoboANKLE angle change during the mechanical experiments.}
		\label{case_study_res}
	\end{center}
\end{figure}

The variation in the ankle angle while applying a load that generates \(130 \, \text{Nm}\) torque at the ankle joint is shown in Fig. \ref{case_study_res}. The experimental results indicate that the prosthesis is able to balance a torque of \(130 \, \text{Nm}\) at ankle dorsi-flexison of \(9.55^\circ\). Experiment results shows that RoboANKLE is 98\% compatible with the maximum dorsi-flexison value at normal walking speed.

\subsection{Able-bodied Experiments}

\begin{figure}[htbp]
	\begin{center}
		\includegraphics[width=0.5\linewidth]{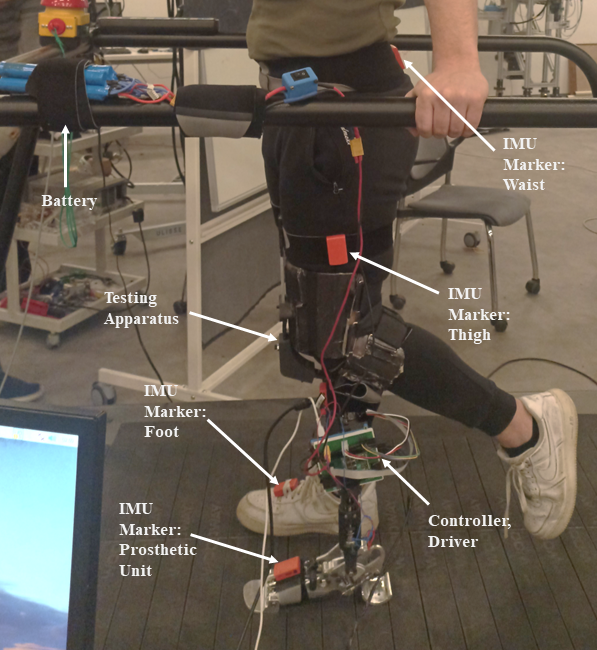}
		\caption{Experimental evaluation environment of the RoboANKLE,  the IMUs placement on the lower-limb, and the apparatus}
		\label{test_setup}
	\end{center}
\end{figure}

To ensure the safety and evaluate the initial functionality of the RoboANKLE, preliminary experiments are conducted with a able-bodied participant, represented by one of the authors of this study. The mass and height of the participant are 75~kg and 173~cm.
The functional evaluations of the RoboANKLE have been approved by the Clinical Research Ethics Committee of Istanbul Medipol University under document number E-10840098-772.02-7438. The participants in this study confirmed their participation by signing the voluntary consent form. Testing apparatus is designed to functionally evaluate the RoboANKLE prototype where it allows the able-bodied participant to wear the prototype for simulating walking motion for experimental evaluation and fixes the flexion angle of the knee joint at 40\(^{\circ}\). XSENS (MVN XSENS, Netherlands) is used to measure the kinematics of the lower-body during the experiments.
Seven XSENS IMUs are placed on the participant limbs and the prototype. These IMUs are placed on the body segments from top the bottom respectively, pelvis, thigh, leg (shank), and foot. The leg and foot IMUs for the prosthesis side are placed on testing apparatus and the prototype. The duration of the experiment is 60 s and the data are captured at sampling rate of 100 Hz. After finishing the experiments, the collected data are post processed with Xsens MVN application and Euler rotation angles for each joint are exported. 
The adjustable treadmill (Woodway's PPS MED) is utilized during the experiments. The treadmill speed is set to be as the normal walking speed (1.3 m/s). The results of these experiments are compared with kinematic and kinetic results of the normal walking speed in the literature \cite{Winter2009}. Due to the difficulty of walking at higher speeds with the apparatus attached below the knee, the experiments are performed at 1.3 m/s instead of 1.6 m/s at which the calculations regarding the prosthesis are conducted. Additionally, since only working principle and functionality of the proposed mechanisms and the prototype are aimed to be assessed, spreading the torque generation over the time feature which activates by the replacer motor is not used in this step. Hence, the DF mechanism is snapping at the push-off phase to generate the required torque around the ankle joint. Experimental setup with IMUs positioning is shown in Fig. \ref{test_setup}.

\begin{figure}[htbp]
	\begin{center}
		\includegraphics[width=\textwidth]{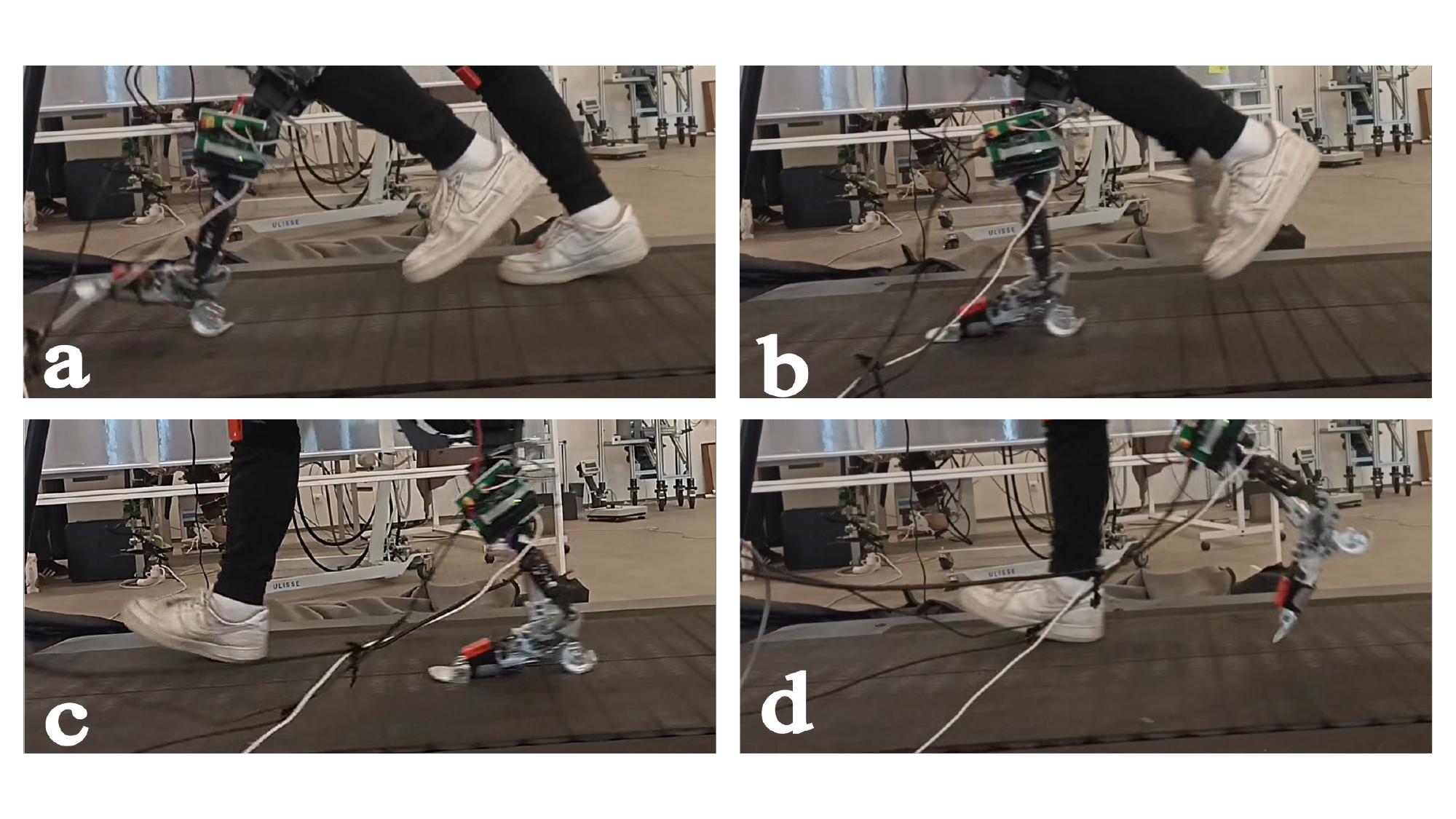}
		\caption{The frames of the walking with RoboANKLE at normal speed,  Heel-strike (a), midstance (b), push-off - dorsi at maximum ankle angle (c), and push-off - plantar-flexion at maximum ankle (d)}
		\label{testing_frames}
	\end{center}
\end{figure}

In Fig. \ref{testing_frames}, a walking cycle of the RoboANKLE is depicted. In (a), the cycle starts with the heel-strike phase, where the heel strike occurs, and no torque is applied to the ankle joint by the DF and EES mechanisms. At this moment, the EES mechanism stores energy. In (b), as the prototype transitions into mid-stance, the DF mechanism begins to store energy, and the EES mechanism continues its energy storage. (c) illustrates the onset of the push-off phase, during which the foot achieves maximum dorsi-flexion. The DF mechanism initiates the release of stored energy into the system, generating torque around the ankle joint. Simultaneously, the EES mechanism is activated as its lock disengages, snapping into the required position and contributing to torque generation at the ankle joint. This activation in EES mechanism requires minimal energy expenditure from its stored energy. Finally, in (d), the push-off phase concludes with the foot reaches to maximum plantar-flexion. All generated torque is conveyed to the prototype's ankle joint between (c) and (d). After the toe lifts, the swing phase begins. Due to the maximum plantar-flexion motion, the locks of the EES mechanism are re-engaged. From there walking cycle finishes and a new cycle starts with heel strike. 

\begin{figure*}[h]
	\centering
	\hspace{-8mm}
	\subfloat[]{
		\includegraphics[width=0.27\textwidth]{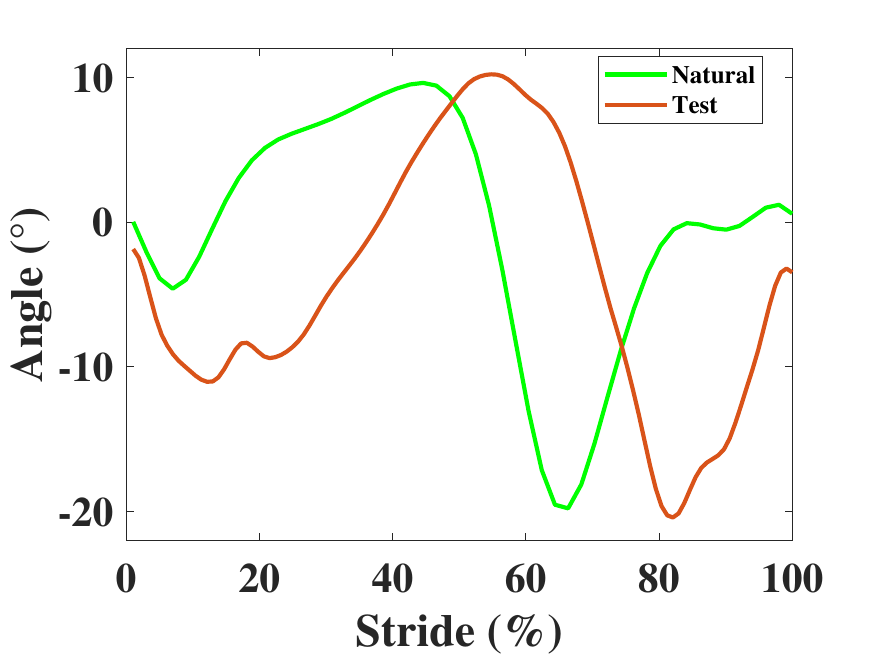}
		\label{subfig:result_angle}
	}
	\hspace{-8mm} 
	\subfloat[\scriptsize]{
		\includegraphics[width=0.27\textwidth]{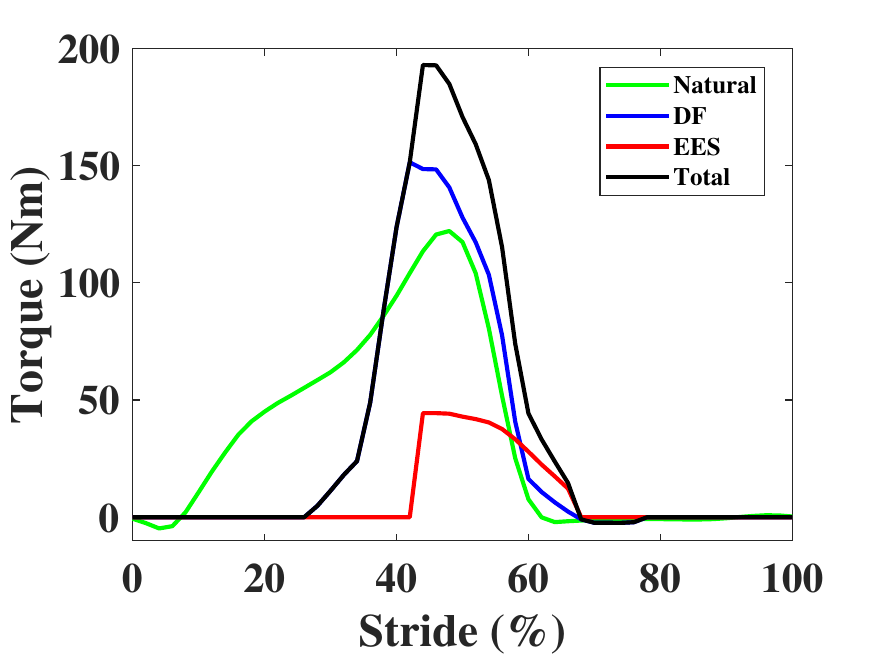}
		\label{subfig:result_torque_ang}
	}
	\hspace{-8mm}
	\subfloat[\scriptsize]{
		\includegraphics[width=0.27\textwidth]{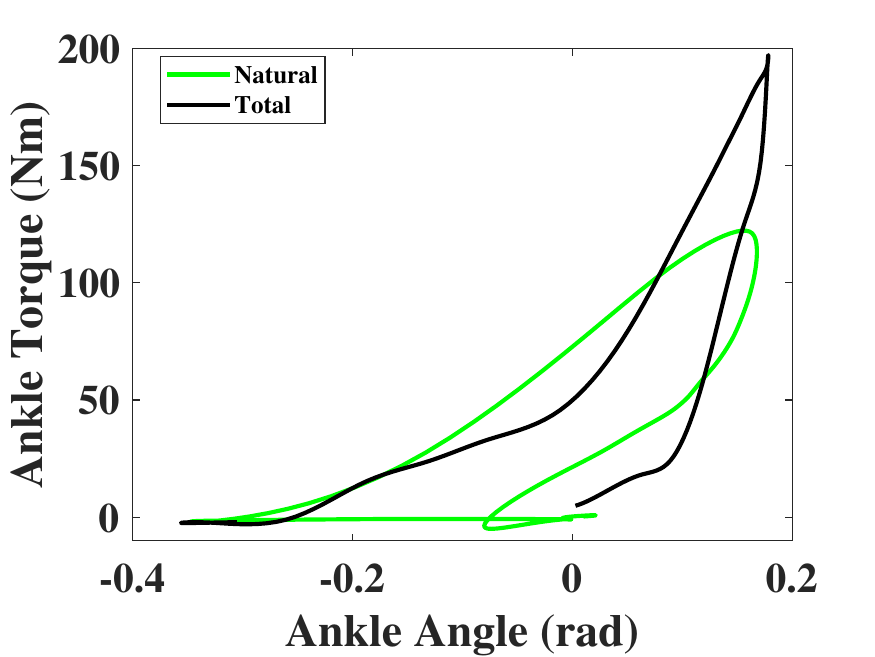}
		\label{subfig:result_torque_rad}
	}
	\hspace{-8mm} 
	\subfloat[\scriptsize]{
		\includegraphics[width=0.27\textwidth]{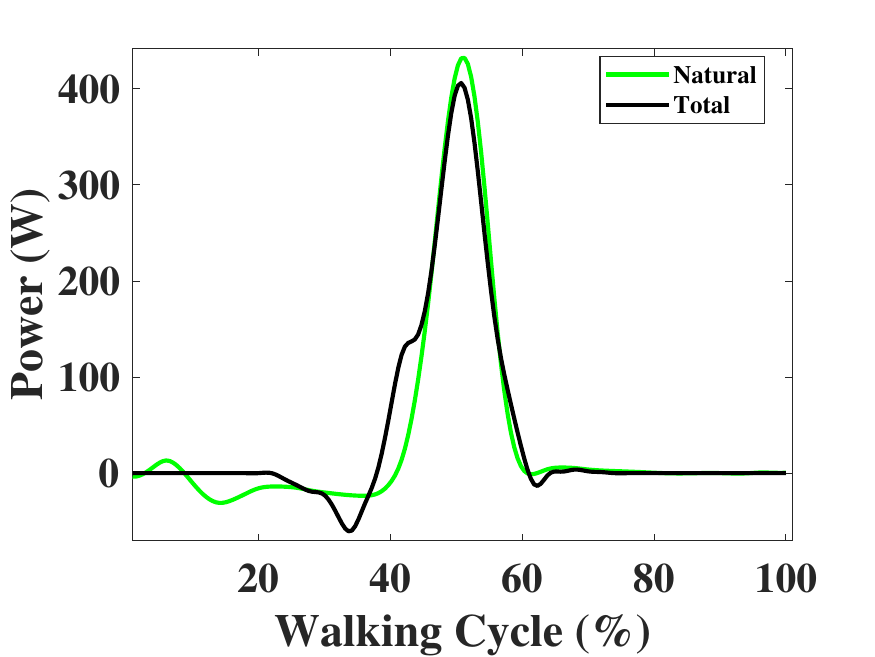} 
		\label{subfig:resultpower}
	}
	
	\caption{Comparison of the natural ankle angle (green) with prototype (black) during walking cycle (a), comparison natural ankle torque (green) with DF generated torque (blue), EES generated torque (red), and total torque generated by prototype (black)(b), comparison of the ankle unit angle and torque between the natural ankle (green) and prototype (black) (c), and comparison of the natural ankle power (green) with prototype (black) during walking cycle (d)}
	\label{fig:ankle_eval}
\end{figure*}

Fig. \ref{subfig:result_angle} illustrates the ankle angles of the prototype and natural ankle movements at normal walking speed during the gait cycle. The prototype's ankle angles represent the average of four consecutive steps in the experiments. The figure depicts the natural ankle angle with a green line, while the prosthetic angle is shown in red. The x-axis denotes the percentage of the gait cycle, and the y-axis represents the ankle angle. The natural ankle initiates with a heel strike, descending to $-4.6^\circ$ in plantar-flexion. Subsequently, dorsi-flexion commences at $13\%$ of the gait cycle, peaking at $9.62^\circ$ at $45\%$. The push-off phase extends from $45\%$ to $66\%$, where the ankle angle reaches a maximum plantar-flexion of $-19.77^\circ$. From $66\%$ to $84\%$, the foot stabilizes to a neutral angle, concluding the gait cycle with a heel strike. For the prosthesis, the heel strike marks the start of the cadence, dropping to $-11.02^\circ$. Dorsiflexion begins at $37\%$ of the cycle, peaking at $10.21^\circ$ at $54\%$. There is an error of 0.56\% . The push-off phase lasts from $54\%$ to $82\%$, culminating in a maximum  of $-20.42^\circ$ at $82\%$. Post-push-off, in the swing phase the prosthesis stabilizes to $-3.2^\circ$, leading up to the heel strike.

Fig. \ref{subfig:result_torque_ang} illustrates the natural ankle torque comparing the experimented torque of the prosthesis and stride. The natural ankle torque starts to build up at 8\% for normal-speed walking and reaches a peak value at 45\%, which later decreases. Torque release occurs from 45\% to 65\% in the push-off phase. As in the prototype, two different mechanisms generate torque. Most of the torque is provided by the DF mechanism. DF mechanism torque generation starts at 26\% of the stride and increases up to 44\% of the stride from 0 Nm to 167.39 Nm. From 42\% to 70\%, torque release occurs in the prototype. In the EES mechanism, torque build-up starts at 44\% when the dorsi-flexion is at maximum, and the push-off starts. Reaches up to 44 Nm and decreases to 0 Nm until the push-off finishes at 70\% of the stride. The total torque provided by the prototype starts building up at 26\% and reaches its maximum at 44\% of 209 $Nm$ at dorsi-flexion maximum. Moreover, the release ends at 70\%. The prototype's torque generation is 58\% higher than the natural ankle torque. 

Fig. \ref{subfig:result_torque_rad} illustrates ankle torque and angle graph. In the figure, "red" depicts the natural ankle torque, "green" represents the DF mechanism torque, "blue" represents the EES mechanism torque, and "black" represents the total, which means the EES and DF mechanism's added torque. This figure depicts the relation between ankle joint rotation and ankle joint torque. In the natural ankle, torque peaks at 0.16 rad, and the torque storing happens at approximately 0.22 rad and this torque is fully released at -0.34 rad at maximum plantar-flexion. Rather than that, in the DF mechanism, the torque builds up to 0.17 rad and releases until -0.34 rad. The EES mechanism torque releases from maximum dorsi-flexion until the end of push-off, which is between 0.177 rad and -0.34 rad. 

Fig. \ref{subfig:resultpower} illustrates the power graph of the prototype and natural ankle. The natural ankle power starts increasing from 37\% of the stride and reaches a peak at 52\% with a value of 432 Watts. Natural ankle minimum power is -30 Watts. In the prototype, the total power provided peaks at a maximum of 401 Watts. The natural ankle has 1900 Joule of energy. On the other hand, the prototype provides 2104.5 Joule.

\section{Discussion}

In this study, design, development, and functional evaluations of a robotic ankle with a motorized compliant unit is proposed. The design of the RoboANKLE employs two torque-generation unit. While the the first mechanism (DF) is responsible for storing energy in theDF spring starting from the mid-stance phase and return it to the ankle unit at the push-off phase, the EES mechanism stores energy throughout the gait cycle except the push-off phase and release it at the push-off phase. The prototype of the RoboANKLE has a weight of 1.92 kg and its dimensions are $26\times107\times420 mm$. The proposed prototype is 30\% lighter than the low-power ankle-foot prosthesis \cite{mazzarini2023} and as light as the powered prosthesis proposed in \cite{liu2024design}. The characterization results of the RoboANKLE shows that, this prototype is capable of generating 7.7\% more torque than the required torque for the fast walking and is 75\% faster than the required speed for compression the motorized spring during the gait cycle. The results of the velocity control for the motorized spring show that, the control method is works properly where the correlation coefficient between the reference and the measured values are 97.8\%. Mechanical experiments of the RoboANKLE reveals that the DF mechanism is able to reach the 98\% of the maximum dorsiflexion angle for the normal walking values. The functional evaluations are performed with an able-bodied to assess the functionality of the RoboANKLE. The results of these experiments shows that, the RoboANkle is capable of generating 57\% higher torque than the required for the normal walking. Additionally, the results show that, the power generation capacity of prototype is 10\% more than the required value for the gait cycle at normal speed.

Compared to the natural ankle angle, there are differences between the measured ankle angles of the RoboANKLE. Although the original aim is to mimic the natural ankle's movement with mechanisms for mimicking muscular structures and managing power during motion mechanics, some differences occurred. These arise during the RoboANKLE experiments with the proposed testing apparatus, which alters walking patterns. Discomfort from wearing the testing apparatus, causing pain over the kneecap, also made it challenging to walk smoothly. Even though the goal is not to precisely replicate the ankle angle profile but to evaluate functionality, the prototype surpassed the natural ankle angle by 5\% ($10.22^{\circ}$) in dorsi-flexion. The dorsi-flexion from a neutral state is comparable between the RoboANKLE and the natural ankle angle. After the heel-strike, dorsi-flexion in the natural ankle starts immediately. It continues until the maximum value, whereas in the prototype, after the heel strike, the foot rotates toward plantar-flexion to a greater degree than the natural ankle, affecting the transition back to dorsi-flexion. This results in the dorsi-flexion towards the maximum value, starting later than in the natural ankle. The dorsi-flexion occurs earlier in the natural ankle than in the RoboANKLE, and the push-off takes 27\% longer to occur.
The maximum plantar-flexion is 3\% higher than in the natural ankle. This demonstrates that the prototype can functionally follow the natural ankle and meet the functionality requirements. At heel strike, the prototype reaches up to -10° upon ground contact, which is minimized by a balancing spring mechanism. After push-off, the same balancing spring rotates the foot back to its original position. The prototype rotates to -5° after push-off because the DF spring system is preloaded, and the spring starts at $-5^{\circ}$, which the balancing spring can rotate up to. 

The torque values of the natural ankle over the prototype. The prototype torque values resulted as 57\% higher than the natural ankle torque. The difference occurs because the prototype is designed for fast walking, where fast walking ankle dorsi-flexion is up to 8\textdegree. That over-rotation causes compression of the spring further than intended. Also, when the push-off starts at the maximum dorsi-flexion, the EES mechanism injects torque into a system that leads to a high torque value at the defined point. It seems the torque at the natural ankle starts building up at 10\% while for the RoboANKLE, compression starts at 26\%. This is due to delayed dorsi-flexion in the RoboANKLE. In the natural ankle, torque generates after the heel strike, while in the RoboANKLE, the torque generates when the dorsi-flexion occurs. Therefore, the torque generation has a delay compared to the natural ankle. The torque is stored within the dorsi-flexion and released in the push-off phase. The prototype is able to inject torque in a prolonged fashion, comparing building time with 33\%. 

The power generation by the RoboANKLE prosthesis is similar with the natural ankle power profile. The correlation coefficient between the natural and the total RoboANKLE power graph is 97\%. The RoboANKLE is capable of generating 10\% more energy than the required for the gait cycle. The error between the RoboANKLE and the natural ankle in the angle profile is only 5\%.

\section{Conclusion}
In this study a novel design for a robotic ankle with a motorized compliant unit that is capable of producing the desired torque and power of the desired tasks is proposed. The results of the characterization, mechanical, and functional evaluations with an able-bodied participant, show that, thanks to the employed mechanisms in the RoboANKLE, it is capable of performing the task repeatedly and periodically, while meeting the requirements of the study. The topology optimization of the RoboANKLE components, makes it light and strong enough to be used in the functional evaluations. The functional evaluations of the RoboANKLE show that, the prototype is able to generate 57\% and 10\% more than the required torque and power for the gait cycle at normal speed. The difference between the maximum dorsi-flexion angles of the natural ankle and the RoboANKLE prototype during the gait cycle is only 5\%. The functional evaluations with the real users will be performed in the future works. Additionally to reduce the weight of the RoboANKLE further, the components of it will be manufactured from the composite material to maintain both lightness and strength. Furthermore to investigate the stability of the robotic ankle, a second DOF will be added to the prototype to facilitate walking on different grounds such as soft and uneven grounds.

\bibliography{mybib}
\bibliographystyle{ieeetr}
\end{document}